\documentclass{article}

\usepackage{mathtools}
\usepackage{alltt}
\usepackage{booktabs}
\usepackage{enumitem}
\usepackage{graphicx}
\usepackage{listings}
\usepackage{mathpartir}
\usepackage{latexsym}
\usepackage{amssymb}
\usepackage{tikz}
\usepackage{tabularx}
\usepackage{pgfplots}
\usepackage{xspace}
\usepackage{xcolor}
\usepackage{hyperref}
\usepackage{cleveref}
\usepackage{tikz}
\usetikzlibrary{positioning,arrows.meta,shapes.geometric}

\usetikzlibrary{tikzmark,arrows.meta}

\newcommand{\phead}[2]{\makebox[\linewidth][c]{\tikzmarknode{#1}{#2}}}

\newcommand{\emptytarget}[1]{\makebox[\linewidth][c]{\tikzmarknode{#1}{\strut}}}

\renewcommand{\mit}[1]{\mathit{#1}}
\newcommand{\mtt}[1]{\texttt{\small #1}}

\newcommand{\jsnull}{\ensuremath{\mtt{null}}}
\newcommand{\jsstr}[1]{\ensuremath{\texttt{"}{#1}\texttt{"}}}
\newcommand{\jsobj}[1]{\ensuremath{\texttt{\{}{#1}\texttt{\}}}}
\newcommand{\jsarr}[1]{\ensuremath{\texttt{[}{#1}\texttt{]}}}

\newcommand{\jinja}[1]{\ensuremath{\mtt{\textdollar}\texttt{\{}{#1}\texttt{\}}}}

\newcommand{\pdlmodel}[2]{\ensuremath{{\mtt{model:}{#1}\mtt{,}\mtt{input:}{#2}}}}
\newcommand{\pdlcode}[2]{\ensuremath{{\mtt{code:}{#1}\mtt{,}\mtt{lang:}{#2}}}}

\newcommand{\pdldata}[1]{\ensuremath{{\mtt{data:}{#1}}}}
\newcommand{\pdlsequence}[3][]{\ensuremath{{\mtt{sequence:}{#2}\mtt{,} \mtt{join:}{#3}^{#1}}}}

\newcommand{\pdlif}[3]{\ensuremath{{\mtt{if:}{#1}\mtt{,} \mtt{then:}{#2}\mtt{,} \mtt{else:}{#3}}}}

\newcommand{\pdlwhile}[4][]{\ensuremath{{\mtt{while:}{#2}\mtt{,} \mtt{repeat:}{#3}\mtt{,} \mtt{join:}{#4}^{#1}}}}

\newcommand{\pdlfunction}[2]{\ensuremath{{\mtt{function:}{#1}\mtt{,}  \mtt{return:}{#2}}}}
\newcommand{\pdlclosure}[3]{\ensuremath{{\mtt{function:}{#1}\mtt{,}  \mtt{return:}{#2}, \mtt{scope:}{#3}}}}
\newcommand{\pdlcall}[2]{\ensuremath{{\mtt{call:}{#1}\mtt{,} \mtt{args:}{#2}}}}
\newcommand{\pdlfactor}[1]{\ensuremath{{\mtt{factor:}{#1}}}}

\newcommand{\pdlcontext}{\ensuremath{\mtt{pdl\_context}}\xspace}
\newcommand{\pdlscore}{\ensuremath{\mtt{pdl\_score}}\xspace}

\newcommand{\pdlsemd}[3]{\ensuremath{{#2} / {#1} \leadsto {#3}}}
\newcommand{\pdlsemsmc}[2]{\ensuremath{{#1} \rightsquigarrow {#2}}}
\newcommand{\pdlsemsmcstar}[2]{\ensuremath{{#1} \rightsquigarrow^* {#2}}}

\newcommand{\pdlsemp}[4]{\ensuremath{{#2} / {#1} \Rightarrow {#4} / {#3}}}
\newcommand{\pdlsempstar}[4]{\ensuremath{{#2} / {#1} \Rightarrow^* {#4} / {#3}}}
\newcommand{\pdlsembb}[4]{\ensuremath{{#2} / {#1} \rightarrow {#4} / {#3}}}
\newcommand{\pdlsemdefs}[4]{\ensuremath{{#2} / {#1} \xrightarrow{\mit{defs}} {#4} / {#3}}}
\newcommand{\pdlsemparse}[3]{\ensuremath{{#1}, {#2} \xrightarrow{\mit{parse}} {#3}}}
\newcommand{\pdlsemcontrib}[4]{\ensuremath{{#1}, {#2}, {#3} \xrightarrow{\mit{contrib}} {#4}}}

\newcommand{\pdlseme}[3]{\ensuremath{{#1} \vdash {#2} \Downarrow {#3}}}
\newcommand{\pdlseml}[4]{\ensuremath{{#2} \vdash {#3} \Downarrow^{#1} {#4}}}
\newcommand{\pdlllm}[3]{\ensuremath{\mit{sample}({#1}, {#2}) = {#3}}}

 \usepackage[accepted]{icml2026}
\newcommand*{\papertitle}{PPDL: LLM-Based Flows as Probabilistic Programs}
\icmltitlerunning{\papertitle}

\definecolor{keyword}{HTML}{37AC4A} 
\definecolor{jinja2}{HTML}{0070C0}
\definecolor{comment}{HTML}{7F7F7F}
\definecolor{generated}{HTML}{37AC4A} 
\definecolor{toolout}{HTML}{A51DFF}

\lstdefinelanguage{pdl}{
numbers=left,
  numbersep=2mm,
  numberstyle=\tiny\color{gray},
basewidth=0.5em,
  alsoletter={:},
  morekeywords={args:,array:,as:,call:,case:,code:,content:,contribute:,data:,def:,defs:,description:,else:,factor:,fallback:,for:,function:,if:,import:,index:,input:,join:,lang:,lastOf:,match:,maxIterations:,message:,model:,multiline:,num_iterations:,object:,parameters:,parser:,pdl_context:,read:,regex:,repeat:,return:,role:,spec:,text:,then:,type:,until:,with:},
  keywordstyle=\color{keyword}\bf\ttfamily,
  morestring=[s]{$\{}{\}},
  stringstyle=\color{jinja2},
  emphstyle=\color{jinja2},
  showstringspaces=false,
  morecomment=[l]{\#},
  commentstyle=\color{comment}\it,
}

\colorlet{probhig}{gray!70}
\colorlet{probmed}{gray!35}
\colorlet{problow}{white}
\newcommand*{\probcircle}[1]{\tikz[baseline=-0.7ex]\draw[black,fill=#1,radius=0.5em] (0,0) circle ;}

\begin{document}

\twocolumn[
  \icmltitle{\papertitle}
  \begin{icmlauthorlist}
    \icmlauthor{Louis Mandel}{ibm}
    \icmlauthor{Guillaume Baudart}{inria}
    \icmlauthor{Mandana Vaziri}{ibm}
    \icmlauthor{Martin Hirzel}{ibm}
  \end{icmlauthorlist}
  \icmlaffiliation{ibm}{IBM, New York, USA}
  \icmlaffiliation{inria}{Université Paris Cité, Inria, CNRS, IRIF, France}
  \icmlcorrespondingauthor{Louis Mandel}{lmandel@us.ibm.com}
  \icmlkeywords{probabilistic programming, inference scaling}
  \vskip 0.3in
]
\printAffiliationsAndNotice{}

\begin{abstract}
Building reliable applications that leverage large language
models~(LLMs) remains a significant challenge.
While LLMs offer impressive capabilities across diverse tasks, their
outputs often lack accuracy and provide no clear measure of
confidence.
This uncertainty compounds in flows of multiple
calls to LLMs and other tools, making it difficult for developers and
end-users to trust the results.
This paper introduces a probabilistic language for programming LLM-based flows.
It enables developers to quantify and propagate
uncertainty throughout the application's flow, and experiment with different inference scaling techniques
without adding a single line of code beyond the flow's logic.
We present an experimental study to demonstrate this capability, and 
a case study building a theorem proving agent for the Rocq theorem prover.
 \end{abstract}

\newcommand{\gb}[1]{\textcolor{red}{[\textit{GB: #1}]}}

\section{Introduction}\label{sec:introduction}

While a single call to a large language model~(LLM) can perform remarkably
well for some tasks, other tasks require flows of multiple calls to LLMs
and other tools~\cite{jimenez_et_al_2024}.
But while such LLM-based flows may improve accuracy, unfortunately,
they also exacerbate uncertainty.
Because each step depends on previous uncertain outputs,
uncertainty compounds along the trace of a flow's execution.
This
makes it difficult for users to assess the correctness of an
answer, leaving them wondering how certain it is and what other
answers are likely.

\begin{figure}
  \resizebox{\columnwidth}{!}{\definecolor{boxblue}{RGB}{198,222,241}

\begin{tikzpicture}[
    >={Stealth[length=3mm,width=2.5mm]},
    box/.style={
        rectangle,
        draw=black,
        thick,
        fill=boxblue,
        minimum width=4.1cm,
        minimum height=2.7cm,
        align=center,
        font=\LARGE
    },
    node-dark/.style ={circle, draw=black, fill=gray!70, minimum size=6mm, inner sep=0pt},
    node-mid/.style  ={circle, draw=black, fill=gray!35, minimum size=6mm, inner sep=0pt},
    node-light/.style={circle, draw=black, fill=white,   minimum size=6mm, inner sep=0pt},
    arr/.style={->, thick},
    lbl/.style={font=\Large\itshape, align=center},
    sqarr/.style={-{Stealth[length=4mm,width=3.5mm]}, line width=2pt}
]

\node[box] (TL) at (0,0)     {LLM-based\\ flow with\\ factors};
\node[box] (TR) at (8,0)     {LLM-based\\ flow};
\node[box] (BL) at (0,-6)  {distribution\\ of output\\ values};
\node[box] (BR) at (8,-6)  {single output\\ value};

\draw[sqarr] (TL.east) -- (TR.west);
\node[lbl] at (4, 0.45) {ignore factors};

\draw[sqarr] (BL.east) -- (BR.west);
\node[lbl] at (4, -5.5) {pick};

\draw[sqarr] ([xshift=-25]TL.south) -- ([xshift=-25]BL.north);
\node[lbl, anchor=west, align=left, text width=4cm] at (-0.65, -3)
    {probabilistically infer (explore\\trace distribution)};

\draw[sqarr] ([xshift=25]TR.south) -- ([xshift=25]BR.north);
\node[lbl, anchor=east, align=right, text width=4.5cm] at (8.65, -3)
    {classically\\interpret (roll\\out single trace)};

\begin{scope}[shift={(-4.4,0)}]
\node[node-dark]  (a0) at (0.0,  0.0)  {};
    \node[node-mid]   (a1) at (0.0, -1.55) {};
    \node[node-mid]   (a2) at (0.0, -3.10) {};
    \node[node-mid]   (a3) at (0.0, -4.65) {};
\node[node-dark]  (b1) at (0.85,-1.55) {};
    \node[node-dark]  (b2) at (0.85,-3.10) {};
    \node[node-light] (b3) at (0.85,-4.65) {};
\node[node-mid]   (c1) at (1.7, -1.55) {};
    \node[node-light] (c2) at (1.7, -3.10) {};
    \node[node-dark]  (c3) at (1.7, -4.65) {};

\draw[arr] (a0) -- (a1);
    \draw[arr] (a0) -- (b1);
    \draw[arr] (a0) -- (c1);
\draw[arr] (a1) -- (a2);
    \draw[arr] (a2) -- (a3);
\draw[arr] (b1) -- (b2);
    \draw[arr] (b2) -- (b3);
\draw[arr] (c1) -- (c2);
    \draw[arr] (b2) -- (c3);
\end{scope}

\begin{scope}[shift={(10.7,0)}]
    \node[node-dark] (d0) at (0,  0.0)  {};
    \node[node-mid]  (d1) at (0, -1.55) {};
    \node[node-mid]  (d2) at (0, -3.10) {};
    \node[node-mid]  (d3) at (0, -4.65) {};
    \draw[arr] (d0) -- (d1);
    \draw[arr] (d1) -- (d2);
    \draw[arr] (d2) -- (d3);
\end{scope}

\end{tikzpicture} }
\caption{\label{fig:commuting}Executing LLM-based flows with or without probabilistic programming. Each circle represents an execution state, with darker shades indicating higher normalized probability.}
\end{figure}
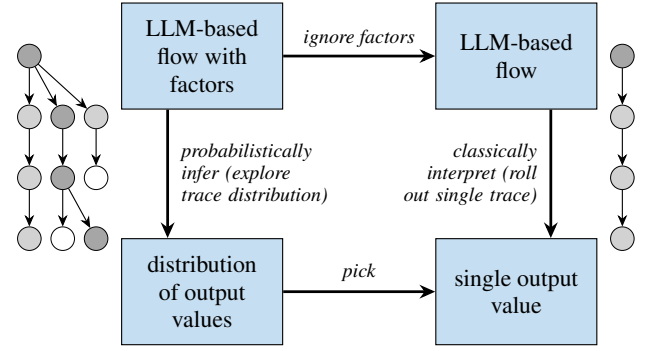

Several recent solutions for improving LLM flow accuracy use
\emph{inference scaling}, which scales up the number of LLM inference
calls or tokens at test time~\cite{chen_et_al_2023,cobbe_et_al_2021,wang_et_al_2023,yao_et_al_2023_tot,yao_et_al_2023_react}.
The idea is to try the same task several times,
either independently or with feedback between attempts, hoping that
one of the outputs is correct.
Inference scaling usually leverages \emph{constraints} to refine or select
outputs, including soft constraints
from models and hard constraints from rule-based tools.
Since it explores a distribution of
outputs and attempts to shape that distribution via constraints,
it could be viewed as a form of probabilistic
programming~\cite{gordon_et_al_2014}.
But in practice, it tends to be more ad hoc.
While inference scaling could save cost by allocating more resources
to better execution traces, that would require a principled way to track
multiple partial traces along with their likelihoods.
However, current inference scaling implementations are tightly
interwoven with the program, increasing complexity and hindering
the exploration of alternative paradigms.
Furthermore, although in theory inference scaling could help users assess
answer quality (since it deals in distributions), in
practice it is rarely principled enough to deliver that benefit.

To obtain more accurate results from LLM-based flows while also giving users
visibility into their uncertainty, we aim to make inference
scaling more principled by incorporating it into a programming language.
We introduce \emph{PPDL}, the first probabilistic programming
language for LLM-based flows.
In general, \emph{Probabilistic programming languages}~\cite{gordon_et_al_2014,carpenter_et_al_2017,goodman_et_al_2008,bingham_et_al_2019}
decouple the search over a distribution of traces from the core
program logic, but they are notoriously hard to use and are not applied to
LLM-based flows.
\emph{Prompt programming languages}~\cite{lundberg_et_al_2022,khattab_et_al_2024,zheng_et_al_2024,mell_et_al_2025}
make it easy to compose flows of 
LLM and tool calls,
but do not help with tracking uncertainty. 
And while they enable developers to explicitly build ad hoc inference 
scaling workflows, such close coupling between the search and the core
logic is complicated and inflexible.

PPDL's goal is to bring these two paradigms together
in a single framework. PPDL is designed for developers and researchers who build multi-step LLM or agentic workflows, including prompt-language users, agent framework developers, and researchers experimenting with inference scaling. PPDL lets users specify the core logic of their program in a
high-level prompting language which already provides one of the two core
constructs of probabilistic languages, namely \emph{sample} in the
form of an LLM call. It extends this language with the second of the two
core constructs of probabilistic languages, namely \emph{factor}.
A factor updates the probability of the current execution trace
based on user-specified soft or hard constraints.

Flow specifications in PPDL require no explicit inference scaling loops, as the runtime takes care of
exploring a distribution of possible traces (see \Cref{fig:commuting}).
Thus, PPDL effectively decouples inference scaling from the core program logic,
providing a suite of plug-and-play probabilistic inference engines: simple majority
voting, importance sampling, and particle
filtering~(aka.\ Sequential Monte Carlo or SMC).
Each particle represents a possible trace through the program,
which might involve multiple LLM and tool calls.
Unlike traditional prompting languages that return a single output,
PPDL returns a distribution of outputs.
Users can then inspect that distribution to assess uncertainty, or
pick a single output, such as the one with the highest probability score.

Besides prompts, sample~(LLM calls), and factor, PPDL includes constructs
for variables, control structures (loops, conditionals, error handling), functions, file imports, and tool calls.
It allows users to write a flow once and then experiment with various
inference scaling engines without writing an additional line of code.
This paper formalizes a semantics for PPDL that clarifies its behavior
as both a prompting language and a probabilistic language, and the
resulting distribution of traces and output values.
Furthermore, this paper describes an implementation with 
parallelization both within a single execution trace and across multiple
in-flight partial traces using functional data structures.
The empirical results show the versatility of PPDL as a
framework for experimenting with different inference scaling algorithms
across multiple widely used benchmarks, as well as a case study where
it is used to implement a theorem proving agent for the Rocq theorem prover~\cite{rocq-prover}.

This paper makes the following novel contributions:

\begin{enumerate}[nosep]
  \item PPDL, the first probabilistic prompt programming language for
    flows of LLM and tool calls.
  \item A semantics that formalizes the interplay between prompt-based
    sample and probabilistic factor.
  \item Results with different probabilistic
    inference engines across different LLMs and benchmarks.
  \item A case study building a theorem proving agent for the Rocq theorem prover.
\end{enumerate}

PPDL is open-source and distributed with PDL: \url{https://github.com/IBM/prompt-declaration-language/tree/icml-26}.

\paragraph{Conflict of Interest Disclosure.}
All the authors, except Guillaume Baudart, are employed by IBM, which leads the development of the Granite models, which were among the ones evaluated in this paper. \section{Overview}\label{sec:overview}

\paragraph{Prompt programming.}
PPDL is based on PDL~\cite{vaziri_et_al_2024}, a declarative approach to prompt programming that combines human readability with ease of execution. PDL programs represent the composition of calls to LLMs and tools, abstracting away the plumbing necessary for such compositions. PDL is based on the premise that interactions with an LLM are mainly for the purpose of generating data. Users specify the shape of data to be generated in YAML, which makes it easy to see properly formatted prompts. PDL adds enough scripting to allow specifying entire flows. 
Code blocks allow users to compose LLM calls with arbitrary code written in Python, Jinja2, or Shell. 
PDL enables implicit accumulation of the context of messages,
freeing users from such concerns, 
and provides type checking and constrained decoding to enforce the shape of LLM outputs. 

Consider the example in \Cref{base}, which illustrates a simple flow: we first ask a model to generate a plan in natural language for how to generate code for a problem, then we chain that with another model call to generate code while taking the plan into account.
Lines~1--5 define some variables for use in the rest of the program.
Line~6 declares the shape of data to be produced as a \lstinline{lastOf}, which means a sequence of blocks whose result is that of the last block in the sequence. Lines~7--20 are a list of blocks comprising the body of \lstinline{lastOf}. 
Lines~7--9 show a prompt asking for an English plan for a code generation problem. Line~10 shows a declarative \lstinline{model} call. It provides the id of the model to call from the variable \lstinline{llm} defined in Line~2~(PDL is based on LiteLLM, so this is a LiteLLM id for a watsonX model).
A PDL program implicitly accumulates the context in the background, so the input to this model call is everything that appeared before. In this case, the input to the model is the prompt on Lines~7--9. Next, we formulate a prompt to generate code (Lines~11--14). This is followed by another \lstinline{model} call on Lines~15--19 for code generation. Its input includes all previous prompts and the response of the first model call. The output is parsed using the regular expression on Line~17, which defines the \lstinline{code} capture group.
On Line~18, a type specification enforces the shape of the output to be a JSON object with a \lstinline{code} field. Line~19 defines the variable \lstinline{solution} to hold the response of the LLM.
Finally, Line~20 contains a Jinja expression that accesses the content of \lstinline{solution.code}. Since the top-level block is a \lstinline{lastOf}, the output of the program is the result of this last block.
The left column of \Cref{sols} shows some possible outputs of the program, since LLM responses are not deterministic.

\begin{figure}[t]
\begin{minipage}[b]{0.8\textwidth}
{
\begin{lstlisting}[emph={llm:,temperature:,problem\_statement:,solution, code:}]
defs:
  llm: watsonx/meta-llama/llama-4-maverick-...
  problem_statement: |
    Write a function to find nth centered hexagonal number.
    assert centered_hexagonal_number(10) == 271
lastOf:
- >
  Generate an English plan for how to generate code for
  the following problem: ${ problem_statement }
- model: ${ llm }
- >
  Generate a complete executable Python function
  definition corresponding to the above plan and problem.
  Generate only a single function definition.
- model: ${ llm }
  parser:
    regex: (.|\n)*```python\n(?P<code>(.|\n)*?)```(.|\n)*
    spec: { code: string }
  def: solution
- ${ solution.code | default("")}
\end{lstlisting}}
\end{minipage}
\caption{\label{base}PDL program generating a Python function.}
\end{figure}

\begin{figure}[t]
\begin{minipage}[b]{0.8\textwidth}
{\footnotesize
\begin{lstlisting}[emph={utils:,llm:,pdl\_llm\_as\_judge:,temperature:,problem\_statement:,constraint:,solution,code:}]
defs:
  llm: watsonx/meta-llama/llama-4-maverick-...
  problem_statement: |
    Write a function to find nth centered hexagonal number.
    assert centered_hexagonal_number(10) == 271  
  utils:
    import: utils.pdl
lastOf:
- >
  Generate an English plan for how to generate code for
  the following problem: ${ problem_statement }
- model: ${ llm }
  def: plan
- defs:
    constraint: |
      This plan for the following problem is correct.
      ${ problem_statement }
  factor: ${ utils.llm_judge(llm, plan, constraint) }
- >
  Generate a complete executable Python function
  definition corresponding to the above plan and problem.
  Generate only a single function definition.
- model: ${ llm }
  parser:
    regex: (.|\n)*```python\n(?P<code>(.|\n)*?)```(.|\n)*
    spec: { code: string }
  def: solution
- factor: ${ utils.score_errors_and_warnings(solution) }
- ${ solution.code | default("")}
\end{lstlisting}}
\end{minipage}
\caption{\label{ppdl}PPDL program adding constraints with \lstinline{factor}.}
\end{figure}

\begin{table}[t]
\caption{\label{sols}(Left) Possible outputs of the PDL code from \Cref{base};
(Right) Posterior probability and particle count added by \Cref{ppdl}.}
\centering
{
\small
\begin{tabular}{@{}lcc@{}}
\toprule
\multicolumn{1}{c}{Output}
& Probability & Count\\
\midrule
\begin{lstlisting}[numbers=none,language=python]
def centered_hexagonal_number(n):
  return 1 + 3 * n * (n + 1)
\end{lstlisting} & $0.00005$ & $3$ \\
\begin{lstlisting}[numbers=none,language=python]
def centered_hexagonal_number(n):
  return 1 + 3 * n * (n - 1)
\end{lstlisting} & $0.99988$ & $1$ \\
\begin{lstlisting}[numbers=none,language=python]
def centered_hexagonal_number(n):
  return 1 + 6 * (n * (n + 1)) // 2
\end{lstlisting} & $0.00007$ & 1 \\
\bottomrule
\end{tabular}
}
\end{table}

\smallskip

\paragraph{Probabilistic prompt programming.}
PPDL extends PDL with a single additional primitive, \lstinline{factor}, but fundamentally changes the execution model by turning the program into a distribution over traces.
In PPDL, the user can use \lstinline{factor} to score executions. The program is then bootstrapped into a probabilistic framework that uses those scores in various algorithms. The output of PPDL is a distribution of results, from which the user can sample the \mbox{answer} with the highest probability, providing a greater likelihood of correctness.

\Cref{ppdl} shows the same program as \Cref{base}, but with added constraints. Lines~6--7 import \lstinline{utils}, a library with utility functions not shown in this figure.
Lines~14--18 score the plan that was generated using an LLM-as-a-judge~\cite{zheng_et_al_2023}.
The function \lstinline{utils.llm_judge} takes a model name, a response, and a constraint, and checks whether the response meets the constraint. It is an LLM call that returns true or false, and we use the log probabilities output by the LLM to compute a score according to the following formula:
\begin{center}
score = $\log\left(\frac{\exp(\textrm{lp}_t)}{\exp(\textrm{lp}_t) + \exp(\textrm{lp}_f)}\right)$
\end{center}
where $\textrm{lp}_t$ and $\textrm{lp}_f$ are the log probabilities for \emph{true} and \emph{false}, respectively.
Line~28 shows a rule-based constraint, where we compute a score based on the errors and warnings from the flake8 linter.

Each execution of the program now generates an output and a score that reflect the quality of the corresponding execution with respect to the user constraints introduced by \lstinline{factor}.
The PPDL runtime then performs \emph{probabilistic inference}, which leverages these scores to estimate the \emph{posterior distribution} of possible outputs.
Probabilistic inference is, in general, intractable, and the distributions computed by LLMs operate over enormous spaces.
PPDL thus relies on Monte Carlo approximate inference algorithms that launch a series of independent executions, called \emph{particles}, to estimate the posterior distribution.
For example, using 5 particles, we obtain the distribution of solutions shown in \Cref{sols}. The solution with high probability is correct, while the other two are incorrect. Three of the five particles found the first solution, so majority voting would have yielded a wrong solution. Appendix~\ref{sec:walkthrough} shows some step-by-step executions of the program.

The advantage of constraints is that they attach probabilities to each response, allowing PPDL to return a correct response even if that response is not part of the majority. In this example, we used importance sampling, but we could have used the same program and selected majority voting or Sequential Monte Carlo. PPDL allows the user to write the logic of the flow once, and experiment with different inference scaling approaches without needing additional code.
Appendix~\ref{sec:alternative} presents alternative implementations of \Cref{base,ppdl} in other frameworks to illustrate the refactoring required when switching inference scaling algorithms.

Inference scaling approaches depend on
\emph{constraints} to steer execution traces and select solutions.
Constraints can be hard~(e.g., failing tests) or soft (e.g., being
scored on a scale), cheap~(e.g., checking syntax) or expensive~(e.g.,
calling a large model).
Sometimes constraints are sound, sometimes complete, but even
when neither sound nor complete, they can improve predictive
performance.
Some inference scaling literature~\cite{cobbe_et_al_2021,stroebl_kapoor_narayanan_2024} refers to
the source of constraints as ``verifiers,'' but we avoid that
terminology because it misleadingly implies soundness.
Other literature, inspired by reinforcement learning, refers to
constraints as \emph{rewards} and distinguishes process rewards (in the
middle of an execution trace) from outcome rewards~(at the end of a completed
trace).
Constraints can come from models, which can be either fine-tuned for
this task~(e.g., process reward models, PRMs) or prompted for this
purpose~\cite{zheng_et_al_2023}.
PPDL supports all of the above forms of constraints via the probabilistic
\lstinline{factor} as a single orthogonal language construct.
 \section{Language and Approach}\label{sec:implementation}

This section presents how PPDL programs execute, formalizing their behavior through a probabilistic semantics.

\paragraph{Overview.}
As shown in \Cref{fig:commuting}, PPDL lets users express an
LLM-based flow as a program with factors (top left), then explores traces of
that program to obtain a distribution of output values (bottom left).
This section formalizes this intuition via a semantics of the form
\mbox{$\pdlsemd{S}{p}{\lambda v.\textrm{Pr}(v)}$},
where $p$ is a program, $S$ is its environment state, $\rightsquigarrow$~means
reduces to, and $\lambda v.\textrm{Pr}(v)$ denotes a categorical
probability distribution over output values~$v$.
To define such a semantics over all traces, we first define a semantics
for a single trace, which has the form
\mbox{$p/S\Rightarrow^* v/S'$}.
We refer to the single-trace semantics as a \emph{weighted sampler},
where the weights come from factors and the sampling happens via LLM
model calls.
The weighted sampler is defined using reduction rules of the form
\mbox{$\pdlsemp{S}{p}{S'}{\mit{p\_or\_v}}$}.
Each reduction rule transforms its left-hand side~(program~$p$ and
environment~$S$) into its right-hand side (the remainder of the
program at the next factor or a final value~$v$, with updated
environment~$S'$).
The reduction $\Rightarrow^*$ is the transitive closure of~$\Rightarrow$
and continues until it produces a value.

During execution, the environment~$S$ keeps track of two special variables:
$S[\pdlcontext]$ accumulates messages, and $S[\pdlscore]$ is the trace likelihood.
LLM calls introduce randomness: each model call samples a response from the distribution defined by an LLM given the current \pdlcontext, resulting in a new message appended to \pdlcontext. 
Each \mtt{factor} increments \pdlscore\ by a value provided as a parameter.
A hard constraint corresponds to a \mtt{factor} with parameter~$-\infty$ (probability~$0$ in log space).
As in classic probabilistic programming languages~\cite{bingham_et_al_2019,carpenter_et_al_2017}, we can assume that a value was sampled from a distribution using a \mtt{factor} whose parameter is the log-density of the distribution at that point.
The value of \pdlscore\ thus represents how well the trace satisfies the user constraints.

\Cref{sec:sampler} formalizes the weighted sampler by defining
the syntax and semantics of core PPDL blocks
(see Appendix~\ref{sec:semantics} for the remaining blocks).
The ideal semantics~($\rightsquigarrow$) uses an infinite weighted sum
over all possible traces~($\Rightarrow^*$).
\Cref{sec:inference} describes the probabilistic inference
engines, each of which computes a finite approximation of the
ideal semantics.
Both the ideal semantics and the approximate inference engines
rely on the weighted sampler.
\Cref{sec:parallel} describes the parallelized implementation of
the weighted sampler and the inference engines.

\subsection{Weighted Sampler}\label{sec:sampler}

\begin{figure}
\small
$$
\begin{array}{rc@{~}l}
    \mit{pdl} & ::= &\phantom{\mid}
      \begin{array}[t]{@{}l@{~}l@{~}r@{}}
      \mtt{\{} 
        & \mtt{defs:} \jsobj{x\mtt{:}\mit{pdl}\mtt{,} \dots\mtt{,} x\mtt{:}\mit{pdl}} \mtt{,}
      \\& \mit{body}\mtt{,}
      \\& \mtt{parser:} \mit{parser}\mtt{,}
      \\& \mtt{contribute:} \mit{contribute}\mtt{,}
      \\& \mtt{def:} \mit{x}
      & \mtt{\}}
      \end{array}
\\

    \mit{body} & ::= &\phantom{\mid}
    \pdldata{\mit{expr}}
    \\ &&\mid
    \pdlcode{\mit{pdl}}{\mit{string}}
    \\ &&\mid
    \pdlmodel{\mit{expr}}{\mit{pdl}}
    \\ &&\mid
    \pdlif{\mit{expr}}{\mit{pdl}}{\mit{pdl}}
    \\ &&\mid
    \pdlsequence{\jsarr{\mit{pdl}\mtt{,} \dots\mtt{,} \mit{pdl}}}{\mit{join}}
    \\ &&\mid
    \pdlwhile{\mit{expr}}{\mit{pdl}}{\mit{join}}
    \\ &&\mid
    \pdlfunction{\mit{types}}{\mit{pdl}}
    \\ &&\mid
    \pdlcall{\mit{expr}}{\mit{expr}}
    \\ &&\mid
    \pdlfactor{\mit{expr}}
\\

    \mit{expr} & ::= &\phantom{\mid}
    \jsnull
    \mid
    \mit{bool}
    \mid
    \mit{number}
    \mid
    \mit{string}
    \mid
    \jinja{\mit{jinja\_expr}}
    \\ &&\mid \jsarr{\mit{expr} \mtt{,} \dots \mtt{,} \mit{expr}}
    \mid \jsobj{x\mtt{:}\mit{expr}\mtt{,} \dots\mtt{,} x\mtt{:}\mit{expr}}
\end{array}
$$
\caption{Syntax of PPDL (using the flow-style of YAML).}
\label{fig:syntax}
\end{figure}

\begin{figure*}
\centering
\small
\begin{mathpar}
\inferrule
{\pdlsemdefs{S}{\mit{defs}}{S_1}{\jsobj{}} \\
 \pdlsembb{S_1}{\mit{block\_body}}{S_2}{v} \\
 \pdlsemparse{\mit{parser}}{v}{v'} \\
 \pdlsemcontrib{S_2}{v'}{\mit{contribute}}{S_3} \\
 S' = S_3[x \leftarrow v']
}
{\pdlsemp{S}{
      \mtt{\{}
      \mtt{defs:} \mit{defs}\mtt{,}
      \mit{block\_body}\mtt{,}
      \mtt{parser:} \mit{parser}\mtt{,}
      \mtt{contribute:} \mit{contribute}\mtt{,}
      \mtt{def:} \mit{x}
      \mtt{\}}
}{S'}{v'}}

\inferrule {\pdlseme{S}{\mit{model}}{m} \\
 \pdlseme{S}{\mit{input}}{i} \\
 \pdlllm{m}{i}{v}
}
{\pdlsembb{S}{\pdlmodel{\mit{model}}{\mit{input}}}{S}{v}}

\inferrule {\pdlseme{S}{e}{w}\\
 S' = S[\pdlscore \leftarrow S[\pdlscore] + w]
}
{\pdlsembb{S}{\pdlfactor{e}}{S'}{\pdldata{\jsstr{}}}}

\inferrule {
 \pdlseme{S}{f}{\pdlclosure{t}{\mit{p}}{S_f}}\\
 \pdlseme{S}{\mit{args}}{\mit{args}'}\\
 args' \in t\\\\
 \pdlsemp{S_f[\pdlcontext \leftarrow S[\pdlcontext], \pdlscore \leftarrow S[\pdlscore]] + \mit{args}'}{p}{S_f'}{v}\\
 S' = S[\pdlcontext \leftarrow S_f'[\pdlcontext],  \pdlscore \leftarrow S_f'[\pdlscore]]
}
{\pdlsembb{S}{\pdlcall{f}{\mit{args}}}{S'}{v}}
\end{mathpar}
\caption{Key reduction rules for blocks~($\pdlsemp{S}{p}{S'}{\mit{p\_or\_v}}$) and block-bodies~($\pdlsembb{S}{\mit{b}}{S'}{\mit{b\_or\_v}}$). The reduction $\pdlseme{S}{e}{v}$ represents the evaluation of an expression~$e$ into a value~$v$ in the environment~$S$. The full semantics is provided in Appendix~\ref{sec:semantics}.}
\label{fig:pdlsem-main}
\end{figure*}

\Cref{fig:syntax} presents the syntax of a simplified kernel of PPDL,
showing its context-free grammar in Backus-Naur Form.
A program is a block comprising a set of variable definitions, the block body (i.e., the instruction to evaluate), a parser to extract the result from the return value of the body, a flag indicating whether the result contributes to the context, and the name of the variable in which to store this result.
The body of a block can be an expression~(\mtt{data}), code for a tool call, an LLM call with its input, a conditional, a sequence of blocks with a join operator to gather result values, a loop with a join operator to gather the results of each iteration, a function definition, a function call, or a factor statement to update the score.
The examples from \Cref{sec:overview} can be compiled into this kernel.
For instance, blocks without an explicit \mtt{defs} clause can be
compiled to use an empty \mtt{defs}, and a \mtt{model} block without
an \mtt{input} clause can be compiled to use \pdlcontext\ as input.

The top rule of \Cref{fig:pdlsem-main} is the main rule defining the weighted sampler reduction \mbox{$\pdlsemp{S}{p}{S'}{\mit{p\_or\_v}}$}.
The notation for the rules in this paper places premises~(sub-steps) above a horizontal
line and conclusions~(the reduction being defined) below the line.
The premises for evaluating a block begin by
evaluating the variable definitions,
then running the block body to produce a value~$v$,
parsing $v$ into a new value~$v'$,
and finally contributing~$v'$ to the background context and
adding it to the environment.
Contribution to the background context is performed by adding a message to the sequence in \pdlcontext~(we omit the role in the semantics for simplicity):

\vspace*{-3mm}
{\small
\begin{mathpar}
\inferrule
{
 \mit{ctx} = S[\pdlcontext] + \jsarr{\jsobj{\mtt{content:} v}}\\
 S' = S[\pdlcontext \leftarrow \mit{ctx}]
}
{\pdlsemcontrib{S}{v}{\jsarr{\pdlcontext}}{S'}}
\end{mathpar}
}

The rest of \Cref{fig:pdlsem-main} shows the key block-body reduction rules.
The \mtt{model} block evaluates the model name and input expressions to get an actual model name~$m$ and a sequence of messages~$i$, then samples the LLM corresponding to~$m$ conditioned on~$i$ to obtain the output.
The \mtt{factor} block computes a score~$w$ and increments the \pdlscore\ variable by this value. It rewrites into a block that returns the empty string.
The rule for a function \mtt{call} illustrates the handling of the special variables \pdlcontext and \pdlscore. As in functional programming languages, it first retrieves the function definition along with its local environment, evaluates the arguments, checks that they are well-typed, then evaluates the function body in the local environment to compute the returned value.
A key property of PPDL is that \pdlcontext and \pdlscore\ are propagated through function calls, allowing accumulation to continue.

Now that we have a weighted sampler, we can define the ideal semantics of a PPDL program as an infinite number of executions of the sampler~(${N \rightarrow \infty}$) and normalize the weights to obtain a categorical distribution:

\vspace*{-3mm}
{\small
\begin{mathpar}
\inferrule
{
 \left\{ \pdlsempstar{S}{p}{S_i}{v_i} \right\}_{1 \le i \le N}\\
\left\{ w_i = \exp(S_i'[\texttt{pdl\_score}]) \right\}_{1 \le i \le N}\\
 W = \textstyle\sum_{1 \le i \le N} w_i
}
{\pdlsemd{S}{p}{\lambda U. \textstyle\sum_{1 \le i \le N}(w_i / W) \times \delta_{v_i}(U)}}
\end{mathpar}
}
\hspace*{-1.5mm}The premise $\left\{ \pdlsempstar{S}{p}{S_i}{v_i} \right\}_{1 \le i \le N}$ denotes a collection of $N$ executions of the program $p$ starting from the same initial environment $S$. Each execution reduces $p$ to a final value $v_i$ and environment~$S_i$. Each $S_i$ contains the variable \pdlscore, which stores the unnormalized weight (in log space) accumulated along the corresponding execution trace. The sum $W$ normalizes the score of each result~$v_i$. The final result is a categorical distribution that, given a set of values~$U$, returns its probability (where $\delta_{v_i}(U)$ is the Dirac distribution that returns $1$ when $v_i \in U$).

\subsection{Inference Engines}\label{sec:inference}

The ideal semantics are intractable as ${N \rightarrow \infty}$, but probabilistic programming inference algorithms provide ways to approximate them. For example, if $N$ is bounded, the rule defining the semantics corresponds to the \emph{Importance Sampling (IS)} algorithm. Each reduction \mbox{$\pdlsempstar{S}{p}{S_i}{v_i}$} is a \emph{particle} computing a value~$v_i$ with its score~$S_i[\pdlscore]$.  
Similarly, majority voting uses the same rule with uniform weights, ignoring~$S_i[\pdlscore]$.

We can also use our weighted sampler to describe the \emph{Sequential Monte Carlo (SMC)} algorithm~\cite{doucet-smc-2006}. With SMC, during execution, particles with a low score drop their current execution path and restart from the state of a particle with a higher score. The state selection is made by sampling from the distribution of program states at each \emph{resampling point}. In PPDL, the resampling points correspond to the \mtt{factor} blocks (i.e., when the reduction $\Rightarrow$ returns a program block~$p$ instead of a value~$v$).

The SMC algorithm is formalized as follows:
{\small
\begin{mathpar}
\inferrule
{
 \pdlsemsmcstar{\delta_{p,S}}{D_{v,S'}}
}
{\pdlsemd{S}{p}{\pi_1(D_{v,S'})}}

\inferrule
{
 \left\{ \pdlsemp{S_i}{p_i}{S_i'}{p_i'} \mid (p_i, S_i) = \mit{sample}(D_{p,S}) \right\}_{1 \le i \le N} \\
 \left\{ w_i = \exp(S_i'[\texttt{pdl\_score}]) \right\}_{1 \le i \le N}\\
 W = \textstyle\sum_{1 \le i \le N} w_i
}
{\pdlsemsmc{D_{p,S}}{\lambda U.\;\textstyle\sum_{1 \le i \le N} (w_i / W) \times \delta_{p'_i, S'_i[{\texttt{pdl\_score} \leftarrow 0}]}(U)}}
\end{mathpar}
}

Starting from a program~$p$ and environment~$S$, we turn this initial state into a Dirac distribution~($\delta_{p,S}$) and then apply the reduction $\pdlsemsmc{D_{p,S}}{D_{p',S'}}$, which rewrites a distribution of program states (pairs of program and environment) into a new distribution. When all programs in the distribution reduce to values~($D_{v,S'}$), the result is the pushforward of the distribution across the first projection~($\pi_1$), yielding the distribution of return values alone.

The reduction~$\pdlsemsmc{D_{p,S}}{D_{p',S'}}$ samples $N$ program/environment pairs from $D_{p,S}$ and applies the weighted sampler~$\Rightarrow$ to compute the next distribution.

\subsection{Parallel Execution}\label{sec:parallel}

PPDL is an interpreted language. Its implementation directly follows the semantics. Both the implementation of the inference engine and of the weighted sampler are parallelized.

Bayesian inference algorithms such as IS and SMC are inherently parallel, since they rely on independent executions of particles. With IS, full traces can be executed in parallel. SMC introduces a synchronization point at each resampling step, when the distribution of intermediate program states is built and sampled. The PPDL interpreter implements this parallelism using multi-threading.

Within the execution of a particle, the PPDL interpreter parallelizes independent model calls using futures~\cite{halstead_1985}.
Model calls return a promise of a response rather than blocking.
When an expression is evaluated, the interpreter traverses the relevant data structures to await the responses of only those model calls on which the expression depends. To limit dependencies, the data structures created by PPDL, especially \pdlcontext, are lazy~\cite{okasaki_1998}, ensuring that only the accessed cell of an array or field of an object is awaited.
 \section{Evaluation}\label{sec:evaluation}

\begin{table}[t]
\caption{Accuracy comparison across benchmarks, models, and inference algorithms. IS@1 is the baseline (single execution), Maj.\ Voting (Majority Voting), IS (Importance Sampling), and SMC (Sequential Monte Carlo) use 5 particles each, and pass@k represents the oracle upper bound. Bold indicates best performance among inference algorithms. All results show mean $\pm$ standard deviation over 3 runs.}
\setlength{\tabcolsep}{3pt}
\centerline{\scriptsize\begin{tabular}{@{}l@{~}c@{~~}c@{~~}c@{~~}c@{~~}c@{}}
\toprule
model & IS@1 & Maj. Voting & IS & SMC & pass@k \\
\midrule
\midrule
\multicolumn{6}{l}{{\bf GSM8k }}\\
\midrule
granite4-small & 83.8\% {\tiny $\pm$ 0.3}\ & 90.4\% {\tiny $\pm$ 0.9}\ & {\bf 93.7\%} {\tiny $\pm$ 0.5}\ & 92.3\% {\tiny $\pm$ 0.3}\ & 96.1\% {\tiny $\pm$ 0.3}\ \\
granite4-micro & 79.5\% {\tiny $\pm$ 0.9}\ & 87.1\% {\tiny $\pm$ 1.6}\ & {\bf 87.8\%} {\tiny $\pm$ 0.4}\ & 84.2\% {\tiny $\pm$ 0.6}\ & 93.7\% {\tiny $\pm$ 0.1}\ \\
llama4-scout & 93.9\% {\tiny $\pm$ 0.2}\ & 94.5\% {\tiny $\pm$ 0.3}\ & 94.4\% {\tiny $\pm$ 0.3}\ & {\bf 94.6\%} {\tiny $\pm$ 0.3}\ & 95.5\% {\tiny $\pm$ 0.2}\ \\
llama4-maverick & 95.3\% {\tiny $\pm$ 0.2}\ & {\bf 95.7\%} {\tiny $\pm$ 0.4}\ & 95.4\% {\tiny $\pm$ 0.3}\ & 95.3\% {\tiny $\pm$ 0.3}\ & 96.7\% {\tiny $\pm$ 0.3}\ \\
gpt-oss-120b & 86.9\% {\tiny $\pm$ 0.3}\ & 90.4\% {\tiny $\pm$ 0.8}\ & {\bf 92.2\%} {\tiny $\pm$ 0.3}\ & 90.8\% {\tiny $\pm$ 0.0}\ & 95.1\% {\tiny $\pm$ 0.1}\ \\
gpt-oss-20b & 90.2\% {\tiny $\pm$ 1.2}\ & {\bf 93.1\%} {\tiny $\pm$ 0.5}\ & 92.7\% {\tiny $\pm$ 0.4}\ & 91.9\% {\tiny $\pm$ 0.8}\ & 95.5\% {\tiny $\pm$ 0.2}\ \\
\midrule
\multicolumn{6}{l}{{\bf Math500 }}\\
\midrule
granite4-small & 60.7\% {\tiny $\pm$ 0.4}\ & 62.1\% {\tiny $\pm$ 0.6}\ & 64.0\% {\tiny $\pm$ 1.1}\ & {\bf 65.5\%} {\tiny $\pm$ 1.7}\ & 75.7\% {\tiny $\pm$ 0.3}\ \\
granite4-micro & 47.4\% {\tiny $\pm$ 0.6}\ & 47.7\% {\tiny $\pm$ 1.2}\ & {\bf 52.5\%} {\tiny $\pm$ 1.1}\ & 51.1\% {\tiny $\pm$ 0.7}\ & 66.5\% {\tiny $\pm$ 0.5}\ \\
llama4-scout & 66.3\% {\tiny $\pm$ 0.1}\ & 65.7\% {\tiny $\pm$ 0.5}\ & 67.3\% {\tiny $\pm$ 0.4}\ & {\bf 67.5\%} {\tiny $\pm$ 0.6}\ & 71.9\% {\tiny $\pm$ 0.3}\ \\
llama4-maverick & 69.9\% {\tiny $\pm$ 0.3}\ & 69.7\% {\tiny $\pm$ 0.6}\ & {\bf 71.5\%} {\tiny $\pm$ 0.9}\ & 70.1\% {\tiny $\pm$ 0.5}\ & 74.7\% {\tiny $\pm$ 0.6}\ \\
gpt-oss-120b & 74.2\% {\tiny $\pm$ 0.2}\ & 73.8\% {\tiny $\pm$ 1.1}\ & {\bf 74.8\%} {\tiny $\pm$ 1.0}\ & 74.5\% {\tiny $\pm$ 0.3}\ & 77.9\% {\tiny $\pm$ 0.3}\ \\
gpt-oss-20b & 74.1\% {\tiny $\pm$ 0.1}\ & 74.0\% {\tiny $\pm$ 1.1}\ & {\bf 75.7\%} {\tiny $\pm$ 0.5}\ & 74.9\% {\tiny $\pm$ 0.4}\ & 79.0\% {\tiny $\pm$ 0.2}\ \\
\midrule
\multicolumn{6}{l}{{\bf MBPP }}\\
\midrule
granite4-small & 69.7\% {\tiny $\pm$ 1.9}\ & 69.7\% {\tiny $\pm$ 2.9}\ & {\bf 80.5\%} {\tiny $\pm$ 2.1}\ & 80.2\% {\tiny $\pm$ 2.0}\ & 92.0\% {\tiny $\pm$ 1.2}\ \\
granite4-micro & 71.9\% {\tiny $\pm$ 1.0}\ & 72.9\% {\tiny $\pm$ 2.5}\ & {\bf 75.1\%} {\tiny $\pm$ 2.6}\ & 74.8\% {\tiny $\pm$ 0.3}\ & 89.5\% {\tiny $\pm$ 1.8}\ \\
llama4-scout & 82.5\% {\tiny $\pm$ 0.9}\ & 81.7\% {\tiny $\pm$ 2.0}\ & {\bf 86.1\%} {\tiny $\pm$ 2.1}\ & 85.4\% {\tiny $\pm$ 1.6}\ & 92.0\% {\tiny $\pm$ 1.6}\ \\
llama4-maverick & 90.3\% {\tiny $\pm$ 0.5}\ & 90.4\% {\tiny $\pm$ 2.3}\ & {\bf 91.4\%} {\tiny $\pm$ 1.0}\ & 90.9\% {\tiny $\pm$ 1.3}\ & 95.9\% {\tiny $\pm$ 0.9}\ \\
gpt-oss-120b & 92.4\% {\tiny $\pm$ 0.2}\ & 93.6\% {\tiny $\pm$ 0.3}\ & {\bf 94.1\%} {\tiny $\pm$ 1.1}\ & 93.4\% {\tiny $\pm$ 1.0}\ & 97.8\% {\tiny $\pm$ 0.3}\ \\
gpt-oss-20b & 92.3\% {\tiny $\pm$ 0.9}\ & {\bf 93.1\%} {\tiny $\pm$ 1.8}\ & 91.7\% {\tiny $\pm$ 1.2}\ & 92.4\% {\tiny $\pm$ 1.0}\ & 98.0\% {\tiny $\pm$ 0.5}\ \\
\midrule
\multicolumn{6}{l}{{\bf LiveCodeBench }}\\
\midrule
granite4-small & 15.8\% {\tiny $\pm$ 0.3}\ & 15.5\% {\tiny $\pm$ 0.8}\ & {\bf 17.4\%} {\tiny $\pm$ 1.1}\ & 15.7\% {\tiny $\pm$ 0.6}\ & 24.0\% {\tiny $\pm$ 0.7}\ \\
granite4-micro & 11.1\% {\tiny $\pm$ 0.3}\ & {\bf 11.5\%} {\tiny $\pm$ 1.3}\ & 10.7\% {\tiny $\pm$ 0.6}\ & 11.1\% {\tiny $\pm$ 0.9}\ & 18.2\% {\tiny $\pm$ 0.7}\ \\
llama4-scout & 23.9\% {\tiny $\pm$ 0.3}\ & 24.5\% {\tiny $\pm$ 0.6}\ & {\bf 24.9\%} {\tiny $\pm$ 0.4}\ & 23.9\% {\tiny $\pm$ 1.0}\ & 30.3\% {\tiny $\pm$ 0.6}\ \\
llama4-maverick & 29.0\% {\tiny $\pm$ 0.2}\ & 28.5\% {\tiny $\pm$ 0.3}\ & 29.1\% {\tiny $\pm$ 0.2}\ & {\bf 29.3\%} {\tiny $\pm$ 0.1}\ & 36.7\% {\tiny $\pm$ 0.2}\ \\
gpt-oss-120b & 30.3\% {\tiny $\pm$ 0.8}\ & 19.5\% {\tiny $\pm$ 1.5}\ & {\bf 36.1\%} {\tiny $\pm$ 2.0}\ & 35.8\% {\tiny $\pm$ 0.8}\ & 48.2\% {\tiny $\pm$ 0.3}\ \\
gpt-oss-20b & 20.8\% {\tiny $\pm$ 0.5}\ & 12.5\% {\tiny $\pm$ 1.2}\ & 8.0\% {\tiny $\pm$ 0.9}\ & {\bf 15.9\%} {\tiny $\pm$ 0.8}\ & 40.2\% {\tiny $\pm$ 0.3}\ \\
\midrule
\multicolumn{6}{l}{{\bf Fever }}\\
\midrule
granite4-small & 76.6\% {\tiny $\pm$ 0.6}\ & 77.5\% {\tiny $\pm$ 0.8}\ & 77.9\% {\tiny $\pm$ 1.0}\ & {\bf 78.1\%} {\tiny $\pm$ 1.3}\ & 88.3\% {\tiny $\pm$ 0.9}\ \\
granite4-micro & 71.7\% {\tiny $\pm$ 0.2}\ & 73.6\% {\tiny $\pm$ 0.3}\ & {\bf 73.7\%} {\tiny $\pm$ 0.8}\ & 73.3\% {\tiny $\pm$ 1.2}\ & 90.3\% {\tiny $\pm$ 1.1}\ \\
llama4-scout & 85.6\% {\tiny $\pm$ 0.4}\ & 85.0\% {\tiny $\pm$ 0.2}\ & {\bf 85.9\%} {\tiny $\pm$ 1.5}\ & 84.1\% {\tiny $\pm$ 0.1}\ & 93.7\% {\tiny $\pm$ 0.8}\ \\
llama4-maverick & 81.6\% {\tiny $\pm$ 0.3}\ & {\bf 81.7\%} {\tiny $\pm$ 0.6}\ & 80.9\% {\tiny $\pm$ 0.4}\ & 81.0\% {\tiny $\pm$ 0.2}\ & 85.7\% {\tiny $\pm$ 0.3}\ \\
gpt-oss-120b & 86.5\% {\tiny $\pm$ 0.2}\ & {\bf 87.5\%} {\tiny $\pm$ 0.5}\ & 86.7\% {\tiny $\pm$ 0.3}\ & 87.2\% {\tiny $\pm$ 0.0}\ & 92.7\% {\tiny $\pm$ 0.6}\ \\
gpt-oss-20b & 79.9\% {\tiny $\pm$ 0.4}\ & {\bf 81.3\%} {\tiny $\pm$ 0.2}\ & 79.9\% {\tiny $\pm$ 1.0}\ & 81.2\% {\tiny $\pm$ 0.5}\ & 90.6\% {\tiny $\pm$ 0.8}\ \\

\bottomrule
\end{tabular}}
\label{tab:results}
\end{table}

\subsection{PPDL as an Inference Scaling Framework}

This section evaluates PPDL as a unified inference-scaling framework across multiple tasks and models. Our goal is not to claim state-of-the-art task performance, but to demonstrate that PPDL enables users to write a single program and experiment with different inference algorithms without rewriting the flow logic. This allows systematic exploration of which inference strategy works best for a given task and model combination.

\paragraph{Models.}
We used granite4-small,
granite4-micro,
gpt-oss-120b,
gpt-oss-20b,
llama4-maverick, and
llama4-scout in our experiments, covering different families and sizes of models.
Appendix~\ref{sec:models} includes their release and cutoff dates.
We set the model temperature for all experiments to 0.8, and use the default reasoning effort for gpt-oss-120b~(medium). All other hyperparameters were left at their default values.

\paragraph{Benchmarks.}
We use benchmarks that span a variety of tasks and cap all datasets at 500 samples:
\begin{itemize}[nosep,leftmargin=1em]
\item \textbf{Reasoning:} GSM8k~\cite{cobbe_et_al_2021} is a dataset of grade-school level math problems; Math500~\cite{math-500} is a curated selection of 500 problems from the Math dataset~\cite{hendrycksmath2021}, spanning algebra, geometry, number theory, pre-calculus, and probability. 
\item \textbf{Code generation:} MBPP~\cite{austin_et_al_2021} is a collection of mostly basic Python programming tasks; LiveCodeBench~\cite{jain2024livecodebenchholisticcontaminationfree} is a dynamic contamination-free benchmark using fresh problems from LeetCode, AtCoder, and Codeforces. We selected the code generation task and its latest version (v6) with a time window from 8/1/2024 to 5/1/2025.
\item \textbf{Question answering:} Fever~\cite{thorne_et_al_2018} is a fact checking dataset.
\end{itemize}

\paragraph{Inference scaling algorithms.}
We consider the following inference scaling algorithms: majority voting~(Maj.\ Voting), importance sampling~(IS), and sequential Monte Carlo~(SMC). We run all experiments with 5 particles.

\paragraph{PPDL Programs.}
For each benchmark, we write a PPDL program to capture the logic of interacting with LLMs and add factors to shift the distribution toward desirable results using both LLM-based judges and rule-based ones.
The complete PPDL implementations are in \Cref{sec:benchmark-code}.
\begin{itemize}[nosep,leftmargin=1em]
\item \textbf{Reasoning:} The program asks the LLM to solve the math problem, then extracts the answer and asks the same LLM to judge the correctness of the solution and score it according to the formula shown in Section~\ref{sec:overview}.
\item \textbf{Code generation:} The program first asks the LLM to generate an English plan for the problem at hand, then uses an LLM to judge the validity of the plan (scoring accordingly). It then generates a Python function corresponding to the plan by calling the model. It extracts the Python code and uses rule-based judges to check that the solution contains function definitions and to evaluate the number of warnings using flake8. Finally, the same LLM is used to judge the correctness of the solution.
\item \textbf{Question answering:} The program first asks the question to the LLM, enabling it to use a Wikipedia search tool. This may result in tool-use requests. We then check whether the topic(s) to be searched are sensible using the same LLM as a judge and call the search tool. We use LLM-as-judge to check whether enough evidence has been gathered and finally call the LLM again to verify or refute the original question given the gathered evidence.
\end{itemize}

\paragraph{Results.}
\Cref{tab:results} reports the results of our experiments. We perform three runs for each task, model, and algorithm and report the mean accuracy and standard deviation. The column labeled IS@1 is the baseline corresponding to a program executed without inference scaling. Similarly, pass@k is an oracle upper bound, where at least one out of $k$ particles has the correct answer.
These benchmarks demonstrate how easily a user can experiment with different inference scaling algorithms for a given task and model and choose between them without writing additional code. This is particularly interesting for SMC, which has a more intricate implementation. The results show the accuracy improvements achieved by inference scaling over the baseline and that while simple majority voting is the best-performing algorithm in some cases (e.g., GSM8k with gpt-oss-20b), in other cases importance sampling or SMC performs best.

In general, importance sampling (IS) and SMC are preferable to majority voting when the \lstinline{factor} statements provide meaningful information, as they use the factors to shift probability mass toward better traces. Between IS and SMC, SMC is particularly advantageous for longer, deeper flows with informative intermediate factors. IS only reweights completed traces, while SMC can resample during execution at factor points, allowing computation to be redirected toward more promising partial traces earlier in the flow. This gives SMC a more meaningful exploration of the search space in multi-step settings, as demonstrated in the theorem-proving case study below.

\subsection{Case Study: Theorem Proving in Rocq}

As a case study, we use PPDL to implement a theorem proving agent for Rocq, formerly known as Coq~\cite{rocq-prover}, a proof assistant, similar to Lean or Isabelle, that can automatically check mathematical proofs. The complete PPDL implementation is provided in \Cref{sec:benchmark-mini-f2f}.

We focus on the MiniF2F dataset~\citep{minif2f,dsp}, a popular benchmark for evaluating theorem proving agents~\citep{atp,copra,magnushammer,poetry}. The test split comprises 244 high-school level math exercises, ranging from simple algebra to International Mathematics Olympiad problems. While the Lean version of this benchmark is already saturated, with state-of-the-art specialized models with aggressive inference scaling reaching up to 99\% accuracy~\cite{hilbert}, the Rocq version~\cite{minif2f-rocq} is more recent, and the contamination risk is minimal. 
The goal of this experiment is not to compare our agent with state-of-the-art solutions, but to measure how much probabilistic inference can improve the performance of a base model on a challenging task.

\paragraph{Experimental setup.}
We adapt an experiment originally designed for Lean to the Rocq prover to study how different prompting and inference scaling strategies affect LLM-based theorem proving (G. Narozniak, personal communication, 2026). A \emph{full proof} generation agent alternates between two steps: LLM-based proof synthesis and prover-based verification. The agent implements a simple proof/repair loop: if the verification step fails, the agent lists all errors and prompts the LLM to fix the proof.

For each exercise, we compare three inference scaling strategies under a limit on the total number of tokens that can be used to find the proof:
1)~IS@1: repeatedly retry until a proof is found or the token budget is exhausted~(this corresponds to sequential scaling);
2)~IS@$k$: run $k$ particles and stop as soon as one of them finds a proof or the token budget is exhausted (this is equivalent to pass@k thanks to the perfect verifier and because scores are ignored when a proof is found);
3)~SMC@$k$: similar to IS, but use the number of errors at each attempt to score the particles and prioritize the most promising proofs.
When a proof is found, the agent terminates the execution of all parallel particles and returns the successful proof.

Using PPDL, we evaluate these three strategies on the same program by adjusting the runtime configuration, varying only the inference algorithm and particle count. We run all experiments with gpt-oss-120b, a temperature of 1.0, and a budget of 1M tokens.

\paragraph{Results.}

\begin{table}[t]
\caption{Number of problems solved out of the 244 MiniF2F-Rocq problems. We present the average $\pm$ one standard deviation over three runs.}
\centerline{
\begin{tabular}{crrr}
\toprule
\multicolumn{1}{c}{Particles} &
\multicolumn{1}{c}{IS} &
\multicolumn{1}{c}{SMC} &
\multicolumn{1}{c}{delta} \\
\midrule
1 & 72.7 $\pm$ 2.9 & \multicolumn{1}{c}{-} & \multicolumn{1}{c}{-} \\
5 & 82.7 $\pm$ 3.9 & 84.7 $\pm$ 1.2 & +2.0 \\
10 & 84.3 $\pm$ 3.9 & 91.3 $\pm$ 1.7 & +7.0 \\
20 & 87.0 $\pm$ 1.6 & 94.3 $\pm$ 2.1 & +7.3 \\
40 & 87.7 $\pm$ 2.4 & 95.0 $\pm$ 2.4 & +7.3 \\
\bottomrule
\end{tabular}}
\label{tab:res_ntp}
\end{table}

\Cref{tab:res_ntp} presents a summary of the results. We observe that with a constant number of tokens the accuracy varies from 30\% (72.7/244 for IS@1) to 39\% (95/244 for SMC@40). With a fixed number of particles, SMC always performs better than IS. 
SMC can resample during execution at factor points, allowing computation to be redirected toward more promising partial traces earlier in the flow.
This gives SMC a more meaningful exploration of the search space in multi-step settings.
SMC also benefits from having more particles since larger particle populations help preserve diversity during resampling and reduce the risk of collapsing too early onto a narrow set of traces.

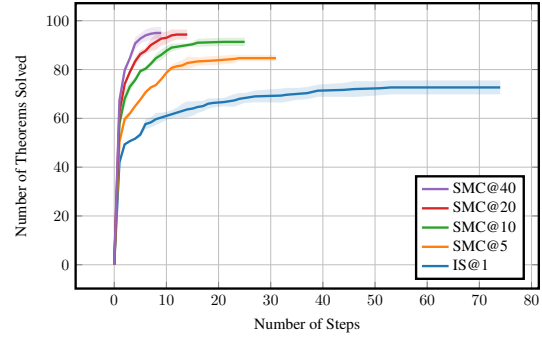
\begin{figure}
\centering
\resizebox{0.925\columnwidth}{!}{\definecolor{plotlyblue}{RGB}{31,119,180}
\definecolor{plotlyorange}{RGB}{255,127,14}
\definecolor{plotlygreen}{RGB}{44,160,44}
\definecolor{plotlyred}{RGB}{214,39,40}
\definecolor{plotlypurple}{RGB}{148,103,189}
\definecolor{plotlybrown}{RGB}{140,86,75}
\definecolor{plotlypink}{RGB}{227,119,194}
\definecolor{plotlygray}{RGB}{127,127,127}
\definecolor{plotlyolive}{RGB}{188,189,34}
\definecolor{plotlycyan}{RGB}{23,190,207}
\begin{tikzpicture}
\begin{axis}[
    xlabel={Number of Steps},
    ylabel={Number of Theorems Solved},
    legend pos=south east,
    legend reversed=true,
    legend cell align={left},
    grid=major,
    width=12cm,
    height=8cm,
    line width=1.5pt,
    mark size=2pt
]

\addplot[color=plotlyblue, fill=plotlyblue, fill opacity=0.15, draw=none, forget plot] coordinates {
    (0,0.0)
    (1,43.41421356237309)
    (2,49.80473785412437)
    (3,51.1380711874577)
    (4,52.1380711874577)
    (5,54.580552462257984)
    (6,59.72147133432299)
    (7,60.38813800098966)
    (8,62.16110492451596)
    (9,62.38813800098966)
    (10,62.63299316185545)
    (11,63.36633983786426)
    (12,64.38813800098966)
    (13,65.94392028877594)
    (14,66.96649831220388)
    (15,66.94392028877594)
    (16,67.16110492451597)
    (17,67.16024689946929)
    (18,68.16024689946929)
    (19,68.21895141649746)
    (21,68.36633983786426)
    (23,69.9580026246706)
    (24,70.44948974278317)
    (25,70.38813800098966)
    (27,71.44948974278317)
    (32,72.2007750890142)
    (33,72.53410842234754)
    (35,72.44948974278317)
    (37,72.82777159118262)
    (39,73.82777159118262)
    (44,74.53410842234754)
    (46,75.2659863237109)
    (51,75.2007750890142)
    (53,75.53410842234754)
    (72,75.53410842234754)
    (73,75.53410842234754)
    (74,75.53410842234754)
    (74,69.7992249109858)
    (73,69.7992249109858)
    (72,69.7992249109858)
    (53,69.7992249109858)
    (51,69.46589157765246)
    (46,68.7340136762891)
    (44,68.7992249109858)
    (39,68.83889507548403)
    (37,67.83889507548403)
    (35,67.55051025721683)
    (33,66.7992249109858)
    (32,66.46589157765246)
    (27,66.55051025721683)
    (25,66.278528665677)
    (24,65.55051025721683)
    (23,64.70866404199606)
    (21,64.96699349546908)
    (19,64.4477152501692)
    (18,63.839753100530714)
    (17,62.839753100530714)
    (16,62.172228408817375)
    (15,61.05607971122405)
    (14,60.36683502112944)
    (13,60.05607971122405)
    (12,60.27852866567701)
    (11,59.96699349546907)
    (10,59.36700683814455)
    (9,58.27852866567701)
    (8,57.17222840881737)
    (7,56.27852866567701)
    (6,55.61186199901034)
    (5,52.08611420440869)
    (4,51.19526214587563)
    (3,50.19526214587563)
    (2,48.8619288125423)
    (1,40.58578643762691)
    (0,0.0)
};

\addplot[color=plotlyblue, mark=none, line width=1.5pt] coordinates {
    (0,0.0)
    (1,42.0)
    (2,49.333333333333336)
    (3,50.666666666666664)
    (4,51.666666666666664)
    (5,53.333333333333336)
    (6,57.666666666666664)
    (7,58.333333333333336)
    (8,59.666666666666664)
    (9,60.333333333333336)
    (10,61.0)
    (11,61.666666666666664)
    (12,62.333333333333336)
    (13,63.0)
    (14,63.666666666666664)
    (15,64.0)
    (16,64.66666666666667)
    (17,65.0)
    (18,66.0)
    (19,66.33333333333333)
    (21,66.66666666666667)
    (23,67.33333333333333)
    (24,68.0)
    (25,68.33333333333333)
    (27,69.0)
    (32,69.33333333333333)
    (33,69.66666666666667)
    (35,70.0)
    (37,70.33333333333333)
    (39,71.33333333333333)
    (44,71.66666666666667)
    (46,72.0)
    (51,72.33333333333333)
    (53,72.66666666666667)
    (72,72.66666666666667)
    (73,72.66666666666667)
    (74,72.66666666666667)
};
\addlegendentry{IS@1}

\addplot[color=plotlyorange, fill=plotlyorange, fill opacity=0.15, draw=none, forget plot] coordinates {
    (0,0.0)
    (1,51.580552462257984)
    (2,61.36633983786426)
    (3,62.816496580927726)
    (4,65.0)
    (5,68.1380711874577)
    (6,71.60947570824874)
    (7,73.1380711874577)
    (8,74.1380711874577)
    (9,76.81649658092772)
    (10,79.1380711874577)
    (11,81.91388579559131)
    (12,83.03300650453092)
    (13,83.721471334323)
    (14,85.29133595800394)
    (15,85.16024689946929)
    (16,85.03300650453092)
    (19,85.5522847498308)
    (23,86.03300650453092)
    (24,85.91388579559131)
    (29,85.91388579559131)
    (31,85.91388579559131)
    (31,83.41944753774203)
    (29,83.41944753774203)
    (24,83.41944753774203)
    (23,82.63366016213574)
    (19,81.78104858350254)
    (16,81.63366016213574)
    (15,80.83975310053071)
    (14,80.0419973753294)
    (13,79.61186199901034)
    (12,79.63366016213574)
    (11,79.41944753774203)
    (10,78.19526214587565)
    (9,75.18350341907228)
    (8,73.19526214587565)
    (7,72.19526214587565)
    (6,69.7238576250846)
    (5,67.19526214587565)
    (4,65.0)
    (3,61.183503419072274)
    (2,57.96699349546907)
    (1,49.08611420440869)
    (0,0.0)
};

\addplot[color=plotlyorange, mark=none, line width=1.5pt] coordinates {
    (0,0.0)
    (1,50.333333333333336)
    (2,59.666666666666664)
    (3,62.0)
    (4,65.0)
    (5,67.66666666666667)
    (6,70.66666666666667)
    (7,72.66666666666667)
    (8,73.66666666666667)
    (9,76.0)
    (10,78.66666666666667)
    (11,80.66666666666667)
    (12,81.33333333333333)
    (13,81.66666666666667)
    (14,82.66666666666667)
    (15,83.0)
    (16,83.33333333333333)
    (19,83.66666666666667)
    (23,84.33333333333333)
    (24,84.66666666666667)
    (29,84.66666666666667)
    (31,84.66666666666667)
};
\addlegendentry{SMC@5}

\addplot[color=plotlygreen, fill=plotlygreen, fill opacity=0.15, draw=none, forget plot] coordinates {
    (0,0.0)
    (1,60.94392028877595)
    (2,71.55902608401044)
    (3,75.16024689946929)
    (4,76.91388579559131)
    (5,80.2761423749154)
    (6,81.2761423749154)
    (7,83.58055246225797)
    (8,86.36633983786426)
    (9,88.16024689946929)
    (10,89.5522847498308)
    (11,90.63299316185545)
    (14,90.81649658092772)
    (15,91.2761423749154)
    (16,92.63299316185545)
    (21,93.03300650453092)
    (22,93.03300650453092)
    (25,93.03300650453092)
    (25,89.63366016213574)
    (22,89.63366016213574)
    (21,89.63366016213574)
    (16,89.36700683814455)
    (15,89.39052429175126)
    (14,89.18350341907228)
    (11,87.36700683814455)
    (10,85.78104858350254)
    (9,83.83975310053071)
    (8,82.96699349546908)
    (7,81.08611420440869)
    (6,79.39052429175126)
    (5,78.39052429175126)
    (4,74.41944753774203)
    (3,70.83975310053071)
    (2,64.44097391598956)
    (1,55.05607971122405)
    (0,0.0)
};

\addplot[color=plotlygreen, mark=none, line width=1.5pt] coordinates {
    (0,0.0)
    (1,58.0)
    (2,68.0)
    (3,73.0)
    (4,75.66666666666667)
    (5,79.33333333333333)
    (6,80.33333333333333)
    (7,82.33333333333333)
    (8,84.66666666666667)
    (9,86.0)
    (10,87.66666666666667)
    (11,89.0)
    (14,90.0)
    (15,90.33333333333333)
    (16,91.0)
    (21,91.33333333333333)
    (22,91.33333333333333)
    (25,91.33333333333333)
};
\addlegendentry{SMC@10}

\addplot[color=plotlyred, fill=plotlyred, fill opacity=0.15, draw=none, forget plot] coordinates {
    (0,0.0)
    (1,65.16110492451595)
    (2,74.81649658092772)
    (3,79.81649658092772)
    (4,85.03300650453092)
    (5,88.03300650453092)
    (6,88.91388579559131)
    (7,92.44948974278317)
    (8,94.63316497887055)
    (9,95.53410842234754)
    (10,95.44948974278317)
    (11,96.44948974278317)
    (12,96.38813800098966)
    (14,96.38813800098966)
    (14,92.278528665677)
    (12,92.278528665677)
    (11,91.55051025721683)
    (10,90.55051025721683)
    (9,89.7992249109858)
    (8,88.0335016877961)
    (7,87.55051025721683)
    (6,86.41944753774203)
    (5,84.63366016213574)
    (4,81.63366016213574)
    (3,78.18350341907228)
    (2,73.18350341907228)
    (1,60.17222840881737)
    (0,0.0)
};

\addplot[color=plotlyred, mark=none, line width=1.5pt] coordinates {
    (0,0.0)
    (1,62.666666666666664)
    (2,74.0)
    (3,79.0)
    (4,83.33333333333333)
    (5,86.33333333333333)
    (6,87.66666666666667)
    (7,90.0)
    (8,91.33333333333333)
    (9,92.66666666666667)
    (10,93.0)
    (11,94.0)
    (12,94.33333333333333)
    (14,94.33333333333333)
};
\addlegendentry{SMC@20}

\addplot[color=plotlypurple, fill=plotlypurple, fill opacity=0.15, draw=none, forget plot] coordinates {
    (0,0.0)
    (1,68.2761423749154)
    (2,81.5522847498308)
    (3,85.1380711874577)
    (4,93.16110492451597)
    (5,94.721471334323)
    (6,96.44948974278317)
    (7,97.16110492451597)
    (8,97.44948974278317)
    (9,97.44948974278317)
    (9,92.55051025721683)
    (8,92.55051025721683)
    (7,92.17222840881738)
    (6,91.55051025721683)
    (5,90.61186199901034)
    (4,88.17222840881738)
    (3,84.19526214587565)
    (2,77.78104858350254)
    (1,66.39052429175126)
    (0,0.0)
};

\addplot[color=plotlypurple, mark=none, line width=1.5pt] coordinates {
    (0,0.0)
    (1,67.33333333333333)
    (2,79.66666666666667)
    (3,84.66666666666667)
    (4,90.66666666666667)
    (5,92.66666666666667)
    (6,94.0)
    (7,94.66666666666667)
    (8,95.0)
    (9,95.0)
};
\addlegendentry{SMC@40}

\end{axis}
\end{tikzpicture} }
\caption{Number of MiniF2F-Rocq problems solved as a function of proof/repair steps. Each curve shows the mean over three runs, with shaded regions indicating one standard deviation.}
\label{fig:res_ntp_smc_steps}
\end{figure}

Since the number of tokens is constant, there is a trade-off between the number of proof/repair steps and multiple particles searching solutions in parallel. \Cref{fig:res_ntp_smc_steps} shows this trade-off for the SMC algorithm (results are similar for IS and shown in \Cref{fig:res_ntp_is_steps} in the appendix).
A single particle~(IS@1) leverages prover feedback across up to 74 sequential repairs whereas SMC@40 goes down to 9 steps.
Parallel exploration is more efficient for exploring the proof space and can prove more theorems in a single step, but it exhausts the token budget more quickly.

Our agent is relatively simple, but optimizing this exploration/exploitation trade-off could be even more beneficial for more advanced prompting strategies that leverage feedback from multiple tools (e.g., the prover, specialized search engines, or a natural language reasoner).
 \section{Related Work}\label{sec:relatedwork}

There is not much prior work at the
intersection between prompt programming and probabilistic programming.
LLaMPPL~\cite{lew_et_al_2023} and GenML
Control~\cite{loula_et_al_2025} apply probabilistic inference engines
to the decoding of LLM-generated tokens with SMC and both soft and hard constraints.
But unlike our work, they
focus on a single LLM call, not a flow of LLM and tool calls.
LMQL features hard constraints and beam
search over traces~\cite{beurerkellner_fischer_vechev_2023};
unlike our work, it lacks soft constraints and probabilistic inference
such as SMC.
Similarly, EnCompass also supports beam
search over traces~\cite{li_et_al_2025}; however, as it does not
track how uncertainty accumulates across factors, it also lacks
probabilistic inference.
Rollout Roulette uses a process reward model with SMC over token
sequences~\cite{puri_et_al_2025}; in contrast, our work supports a
more general set of soft and hard constraints, and supports multi-call flows.
\citet{bertsch_et_al_2023} reframe inference scaling as
probabilistic inference to find the Minimum Bayes Risk sample,
but do not explore SMC nor LLM-and-tool flows.

We take inspiration from the rich literature on probabilistic
languages~(PPLs) that predates prompt programming with LLMs.
An overview paper by \citet{gordon_et_al_2014}
introduces core constructs of probabilistic programming and
inference via the Prob PPL.
The Stan PPL has been widely adopted in practice~\cite{carpenter_et_al_2017}.
The Church PPL is based on lambda calculus~\cite{goodman_et_al_2008}.
Like PPDL, each of Prob, Stan, and Church tracks and updates a log-probability
along a trajectory; but unlike
PPDL, none of them interface with LLMs.
Some more recent PPLs do interface with neural networks, such as
Pyro~\cite{bingham_et_al_2019} and DeepStan~\cite{baudart_et_al_2021_deepstan},
but they do not focus on language models, let alone prompt programming.

Another source of inspiration for PPDL is the growing literature on
prompt programming languages.
We already discussed LMQL and PDL.
Guidance focuses on hard constraints to guide
generation~\cite{lundberg_et_al_2022}.
Both DSPy~\cite{khattab_et_al_2024} and AutoPDL~\cite{spiess_et_al_2025}
provide support for optimizing prompt programs for an objective
function, which could be viewed as a final
soft constraint.
Other prompting languages include SGLang~\cite{zheng_et_al_2024},
Vieira~\cite{li_et_al_2024},
APPL~\cite{dong_et_al_2025},
and Opp~\cite{mell_et_al_2025}.
Unlike PPDL, none of these are probabilistic
languages; while several of the papers include inference scaling
examples, those are written by hand, not orthogonal like in PPDL.

Parallel inference scaling scales the number of independent
attempts for solving a problem~(where each attempt can be a full trace).
Examples include Self-consistency~\cite{wang_et_al_2023} and Large
Language Monkeys~\cite{brown_et_al_2024}.
PPDL makes parallel inference scaling implicit and orthogonal.
Sequential inference scaling scales the number of tokens in a
single LLM call or the number of LLM calls in a single trajectory.
Examples include chain-of-thought~\cite{wei_et_al_2022},
Self-refine~\cite{madaan_et_al_2023}, and
ReAct agents~\cite{yao_et_al_2023_react}.
PDL has been designed to facilitate the implementation of flows that
embody these sequential inference scaling approaches, and
PPDL directly inherits this property.
Besides parallel and sequential, a third style is
search-based inference scaling, e.g.,
\citet{yao_et_al_2023_tot} and \citet{zhou_et_al_2024}.
PPDL makes search-based inference scaling implicit and orthogonal.

Rooted in earlier work on automated machine
learning~\cite{feurer_et_al_2015,baudart_et_al_2021_lale}, automated
prompt optimization uses known question/answer pairs to improve prompts
for future questions.
More recently, \emph{instance optimization} applies prompt
optimization for a specific new question; examples include
TextGrad~\cite{yuksekgonul_et_al_2025},
Trace~\cite{cheng_nie_swaminathan_2024}, and
Gepa~\cite{agrawal_et_al_2025}.
Unlike our work, none are PPLs or use probabilistic inference engines.
In future work, we plan to integrate some of these techniques. \section{Conclusion}\label{sec:conclusion}

This paper introduces PPDL, a new programming language that unifies
prompt programming and probabilistic programming.
Prompt programming languages make LLM prompting easy, but struggle
with inaccuracy and uncertainty.
One solution for boosting accuracy is inference scaling, but
unfortunately, manually adding inference scaling to a prompt
program sacrifices simplicity and flexibility.
Instead, PPDL lets users write the core logic in
a simple prompting language, and the interpreter applies orthogonal
probabilistic inference engines to search the space of trajectories.
As a bonus, since the program returns a distribution,
users also get a better feeling for the uncertainty of results.

\paragraph{Limitations.}

While useful, inference scaling is no silver bullet~\cite{stroebl_kapoor_narayanan_2024}.
Any steering based on factors can only be as good as the information
in those factors, which is often imperfect, e.g., based on an
LLM-as-a-judge~\cite{zheng_et_al_2023}.
In future work, we plan to mitigate that limitation by calibrating factors, so multiple constraints carry the appropriate
relative weight.

\section*{Acknowledgements}

We would like to thank Theo Stoskopf for all the help and support with the rocq-ml-toolbox (\url{https://github.com/LLM4Rocq/rocq-ml-toolbox}) that we used for the theorem-proving case study.

\section*{Impact Statement}

This paper presents work whose goal is to advance the field of Machine Learning.
There are many potential societal consequences of our work, none
of which we feel must be specifically highlighted here.

\clearpage

\bibliographystyle{icml2026}
\bibliography{biblio}

@Misc{agrawal_et_al_2025,
  title = "{GEPA}: Reflective Prompt Evolution Can Outperform Reinforcement Learning",
  author = "Agrawal, Lakshya A and Tan, Shangyin and Soylu, Dilara and Ziems, Noah and Khare, Rishi and Opsahl-Ong, Krista and Singhvi, Arnav and Shandilya, Herumb and Ryan, Michael J and Jiang, Meng and Potts, Christopher and Sen, Koushik and Dimakis, Alexandros G. and Stoica, Ion and Klein, Dan and Zaharia, Matei and Khattab, Omar",
  year = 2025,
  month = jul,
  url = "https://arxiv.org/abs/2507.19457" }

@Misc{austin_et_al_2021,
  title = "Program Synthesis with Large Language Models",
  author = "Austin, Jacob and Odena, Augustus and Nye, Maxwell and Bosma, Maarten and Michalewski, Henryk and Dohan, David and Jiang, Ellen and Cai, Carrie and Terry, Michael and Le, Quoc and Sutton, Charles",
  year = 2021,
  month = aug,
  url = "https://arxiv.org/abs/2108.07732" }

@InProceedings{baudart_et_al_2021_deepstan,
  title = "Compiling {Stan} to Generative Probabilistic Languages and Extension to Deep Probabilistic Programming",
  author = "Baudart, Guillaume and Burroni, Javier and Hirzel, Martin and Mandel, Louis and Shinnar, Avraham",
  booktitle = "Conference on Programming Language Design and Implementation (PLDI)",
  year = 2021,
  month = jun,
  pages = "497--510",
  url = "https://doi.org/10.1145/3453483.3454058" }

@InProceedings{baudart_et_al_2021_lale,
  title = "Pipeline Combinators for Gradual {AutoML}",
  author = "Baudart, Guillaume and Hirzel, Martin and Kate, Kiran and Ram, Parikshit and Shinnar, Avraham and Tsay, Jason",
  booktitle = "Conference on Neural Information Processing Systems (NeurIPS)",
  year = 2021,
  month = dec,
  pages = "19705--19718",
  url = "https://proceedings.neurips.cc/paper/2021/file/a3b36cb25e2e0b93b5f334ffb4e4064e-Paper.pdf" }

@InProceedings{bertsch_et_al_2023,
  title = "It's {MBR} All the Way Down: Modern Generation Techniques Through the Lens of Minimum {Bayes} Risk",
  author = "Bertsch, Amanda and Xie, Alex and Neubig, Graham and Gormley, Matthew R.",
  year = 2023,
  booktitle = "Proceedings of the Big Picture Workshop (BigPicture@EMNLP)",
  month = dec,
  pages = "108--122",
  url = "https://doi.org/10.18653/v1/2023.bigpicture-1.9" }

@InProceedings{beurerkellner_fischer_vechev_2023,
  author = "Beurer-Kellner, Luca and Fischer, Marc and Vechev, Martin",
  title = "Prompting Is Programming: A Query Language for Large Language Models",
  booktitle = "Conference on Programming Language Design and Implementation (PLDI)",
  year = 2023,
  month = jun,
  pages = "1946--1969",
  url = "https://doi.org/10.1145/3591300" }

@Article{bingham_et_al_2019,
  title = "Pyro: Deep Universal Probabilistic Programming",
  author = "Bingham, Eli and Chen, Jonathan P. and Jankowiak, Martin and Obermeyer, Fritz and Pradhan, Neeraj and Karaletsos, Theofanis and Singh, Rohit and Szerlip, Paul and Horsfall, Paul and Goodman, Noah D.",
  journal = "Journal of Machine Learning Research (JMLR)",
  year = 2019,
  volume = 20,
  pages = "1--6",
  url = "https://www.jmlr.org/papers/v20/18-403.html" }

@Misc{brown_et_al_2024,
  title = "Large Language Monkeys: Scaling Inference Compute with Repeated Sampling",
  author = "Brown, Bradley and Juravsky, Jordan and Ehrlich, Ryan and Clark, Ronald and Le, Quoc V. and Re, Christopher and Mirhoseini, Azalia",
  year = 2024,
  month = jul,
  url = "https://arxiv.org/abs/2407.21787" }

@Article{carpenter_et_al_2017,
  author = "Carpenter, Bob and Gelman, Andrew and Hoffman, Matt and Lee, Daniel and Goodrich, Ben and Betancourt, Michael and Brubaker, Michael A. and Guo, Jiqiang and Li, Peter and Riddell, Allen",
  title = "Stan: A probabilistic programming language",
  journal = "Journal of Statistical Software",
  volume = 76,
  number = 1,
  year = 2017,
  pages = "1--37",
  url = "https://www.jstatsoft.org/article/view/v076i01" }

@InProceedings{chen_et_al_2023,
  author = "Chen, Bei and Zhang, Fengji and Nguyen, Anh and Zan, Daoguang and Lin, Zeqi and Lou, Jian-Guang and Chen, Weizhu",
  title = "{CodeT}: Code Generation with Generated Tests",
  booktitle = "International Conference on Learning Representations (ICLR)",
  year = 2023,
  month = may,
  url = "https://openreview.net/forum?id=ktrw68Cmu9c" }

@InProceedings{cheng_nie_swaminathan_2024,
  title = "Trace is the Next {AutoDiff}: Generative Optimization with Rich Feedback, Execution Traces, and {LLMs}",
  author = "Cheng, Ching-An and Nie, Allen and Swaminathan, Adith",
  booktitle = "Conference on Neural Information Processing Systems (NeurIPS)",
  year = 2024,
  month = dec,
  url = "https://proceedings.neurips.cc/paper_files/paper/2024/hash/83ba7056bce2c3c3c27e17397cf3e1f0-Abstract-Conference.html" }

@Misc{cobbe_et_al_2021,
  title = "Training Verifiers to Solve Math Word Problems",
  author = "Cobbe, Karl and Kosaraju, Vineet and Bavarian, Mohammad and Chen, Mark and Jun, Heewoo and Kaiser, Lukasz and Plappert, Matthias and Tworek, Jerry and Hilton, Jacob and Nakano, Reiichiro and Hesse, Christopher and Schulman, John",
  year = 2021,
  month = oct,
  url = "https://arxiv.org/abs/2110.14168" }

@InProceedings{dong_et_al_2025,
  title = "{APPL}: A Prompt Programming Language for Harmonious Integration of Programs and Large Language Model Prompts",
  author = "Dong, Honghua and Su, Qidong and Gao, Yubo and Li, Zhaoyu and Ruan, Yangjun and Pekhimenko, Gennady and Maddison, Chris J. and Si, Xujie",
  booktitle = "Annual Meeting of the Association for Computational Linguistics (ACL)",
  year = 2025,
  month = jul,
  url = "https://aclanthology.org/2025.acl-long.63/" }

@InProceedings{feurer_et_al_2015,
  title = "Efficient and Robust Automated Machine Learning",
  author = "Feurer, Matthias and Klein, Aaron and Eggensperger, Katharina and Springenberg, Jost and Blum, Manuel and Hutter, Frank",
  booktitle = "Conference on Neural Information Processing Systems (NIPS)",
  year = 2015,
  month = dec,
  pages = "2962--2970",
  url = "http://papers.nips.cc/paper/5872-efficient-and-robust-automated-machine-learning" }

@InProceedings{goodman_et_al_2008,
  author = "Goodman, Noah and Mansinghka Vikash and Roy, Daniel and Bonawitz, Keith and Tenenbaum, Joshua",
  title = "Church: a Language for Generative Models",
  booktitle = "Conference on Uncertainty in Artificial Intelligence (UAI)",
  year = 2008,
  pages = "220--229",
  url = "https://dslpitt.org/uai/papers/08/p220-goodman.pdf" }

@InProceedings{gordon_et_al_2014,
  author = "Gordon, Andrew D. and Henzinger, Thomas A. and Nori, Aditya V. and Rajamani, Sriram K.",
  title = "Probabilistic Programming",
  booktitle = "ICSE track on Future of Software Engineering (FOSE@ICSE)",
  year = 2014,
  pages = "167--181",
  url = "https://doi.org/10.1145/2593882.2593900" }

@Article{halstead_1985,
  author = "Halstead, Jr., Robert H.",
  title = "Multilisp: A language for concurrent symbolic computation",
  journal = "Transactions on Programming Languages and Systems (TOPLAS)",
  year = 1985,
  month = oct,
  volume = 7,
  number = 4,
  pages = "501--538",
  url = "https://doi.org/10.1145/4472.4478"}

@InProceedings{jimenez_et_al_2024,
  title = "{SWE-bench}: Can Language Models Resolve Real-World {GitHub} Issues?",
  author = "Jimenez, Carlos E. and Yang, John and Wettig, Alexander and Yao, Shunyu and Pei, Kexin and Press, Ofir and Narasimhan, Karthik",
  booktitle = "International Conference on Learning Representations (ICLR)",
  year = 2024,
  month = may,
  url = "https://openreview.net/forum?id=VTF8yNQM66" }

@InProceedings{khattab_et_al_2024,
  title = "{DSPy}: Compiling Declarative Language Model Calls into Self-Improving Pipelines",
  author = "Khattab, Omar and Singhvi, Arnav and Maheshwari, Paridhi and Zhang, Zhiyuan and Santhanam, Keshav and A, Sri Vardhamanan and Haq, Saiful and Sharma, Ashutosh and Joshi, Thomas T. and Moazam, Hanna and Miller, Heather and Zaharia, Matei and Potts, Christopher",
  booktitle = "International Conference on Learning Representations (ICLR)",
  year = 2024,
  month = may,
  url = "https://openreview.net/forum?id=sY5N0zY5Od" }

@InProceedings{lew_et_al_2023,
  title = "Sequential {Monte} {Carlo} Steering of Large Language Models using Probabilistic Programs",
  author = "Lew, Alexander K. and Zhi-Xuan, Tan and Grand, Gabriel and Mansinghka, Vikash K.",
  booktitle = "Workshop on Sampling and Optimization in Discrete Space (SODS@ICML)",
  year = 2023,
  month = jul,
  url = "https://openreview.net/forum?id=Ul2K0qXxXy" }

@InProceedings{li_et_al_2024,
  title = "Relational Programming with Foundation Models",
  author = "Li, Ziyang and Huang, Jiani and Liu, Jason and Zhu, Felix and Zhao, Eric and Dodds, William and Velingker, Neelay and Alur, Rajeev and Naik, Mayur",
  booktitle = "Conference on Artificial Intelligence (AAAI)",
  year = 2024,
  month = feb,
  url = "https://doi.org/10.1609/aaai.v38i9.28934" }

@InProceedings{li_et_al_2025,
  title = "{EnCompass}: Enhancing Agent Programming with Search Over Program Execution Paths",
  author = "Li, Zhening and Solar-Lezama, Armando and Yue, Yisong and Zheng, Stephan",
  booktitle = "Conference on Neural Information Processing Systems (NeurIPS)",
  year = 2025,
  month = dec,
  url = "https://openreview.net/forum?id=IKVkpjSJzJ" }

@InProceedings{loula_et_al_2025,
  title = "Syntactic and Semantic Control of Large Language Models via Sequential {Monte} {Carlo}",
  author = "Loula, Joao and LeBrun, Benjamin and Du, Li and Lipkin, Ben and Pasti, Clemente and Grand, Gabriel and Liu, Tianyu and Emara, Yahya and Freedman, Marjorie and Eisner, Jason and Cotterell, Ryan and Mansinghka, Vikash and Lew, Alexander K. and Vieira, Tim and O'Donnell, Timothy J.",
  booktitle = "International Conference on Learning Representations (ICLR)",
  year = 2025,
  month = may,
  url = "https://openreview.net/forum?id=xoXn62FzD0" }

@Misc{lundberg_et_al_2022,
  author = "Lundberg, Scott and Ribeiro, Marco Tulio Correia and {et al.}",
  title = "Guidance: A guidance language for controlling large language models",
  year = 2022,
  month = nov,
  url = "https://github.com/guidance-ai/guidance" }

@InProceedings{madaan_et_al_2023,
  title = "Self-Refine: Iterative Refinement with Self-Feedback",
  author = "Madaan, Aman and Tandon, Niket and Gupta, Prakhar and Hallinan, Skyler and Gao, Luyu and Wiegreffe, Sarah and Alon, Uri and Dziri, Nouha and Prabhumoye, Shrimai and Yang, Yiming and Gupta, Shashank and Majumder, Bodhisattwa Prasad and Hermann, Katherine and Welleck, Sean and Yazdanbakhsh, Amir and Clark, Peter",
  booktitle = "Conference on Neural Information Processing Systems (NeurIPS)",
  year = 2023,
  month = dec,
  url = "https://proceedings.neurips.cc/paper_files/paper/2023/hash/91edff07232fb1b55a505a9e9f6c0ff3-Abstract-Conference.html" }

@InProceedings{mell_et_al_2025,
  title = "Opportunistically Parallel Lambda Calculus",
  author = "Mell, Stephen and Kallas, Konstantinos and Zdancewic, Steve and Bastani, Osbert",
  booktitle = "Conference on Object-Oriented Programming, Systems, Languages, and Applications (OOPSLA)",
  year = 2025,
  pages = "2596--2622",
  month = oct,
  url = "https://doi.org/10.1145/3763143" }

@Book{okasaki_1998,
  title = "Purely Functional Data Structures",
  author = "Chris Okasaki",
  publisher = "Cambridge University Press",
  year = 1998 }

@InProceedings{puri_et_al_2025,
  title = "Rollout Roulette: A Probabilistic Inference Approach to Inference-Time Scaling of {LLMs} using Particle-Based {Monte} {Carlo} Methods",
  author = "Puri, Isha and Sudalairaj, Shivchander and Xu, Guangxuan and Xu, Kai and Srivastava, Akash",
  booktitle = "Conference on Neural Information Processing Systems (NeurIPS)",
  year = 2025,
  month = dec,
  url = "https://openreview.net/forum?id=qPQUrjiA0q" }

@InProceedings{spiess_et_al_2025,
  title = "{AutoPDL}: Automatic Prompt Optimization for {LLM} Agents",
  author = "Spiess, Claudio and Vaziri, Mandana and Mandel, Louis and Hirzel, Martin",
  booktitle = "Conference on Automated Machine Learning (AutoML)",
  year = 2025,
  month = sep,
  url = "https://proceedings.mlr.press/v293/spiess25a.html" }

@Misc{stroebl_kapoor_narayanan_2024,
  title = "Inference Scaling {FLaws}: The Limits of {LLM} Resampling with Imperfect Verifiers",
  author = "Stroebl, Benedikt and Kapoor, Sayash and Narayanan, Arvind",
  year = 2026,
  month = mar,
  url = "https://arxiv.org/abs/2411.17501" }

@Misc{thorne_et_al_2018,
  title = "{FEVER}: a large-scale dataset for Fact Extraction and {VERification}",
  author = "Thorne, James and Vlachos, Andreas and Christodoulopoulos, Christos and Mittal, Arpit",
  year = 2018,
  month = mar,
  url = "https://arxiv.org/abs/1803.05355" }

@Misc{vaziri_et_al_2024,
  author = "Vaziri, Mandana and Mandel, Louis and Spiess, Claudio and Hirzel, Martin",
  title = "{PDL}: A Declarative Prompt Programming Language",
  year = 2024,
  month = oct,
  url = "http://arxiv.org/abs/2410.19135" }

@InProceedings{wang_et_al_2023,
  title = "Self-Consistency Improves Chain of Thought Reasoning in Language Models",
  author = "Wang, Xuezhi and Wei, Jason and Schuurmans, Dale and Le, Quoc V and Chi, Ed H. and Narang, Sharan and Chowdhery, Aakanksha and Zhou, Denny",
  booktitle = "International Conference on Learning Representations (ICLR)",
  year = 2023,
  month = may,
  url = "https://openreview.net/forum?id=1PL1NIMMrw" }

@InProceedings{wei_et_al_2022,
  title = "Chain-of-Thought Prompting Elicits Reasoning in Large Language Models",
  author = "Wei, Jason and Wang, Xuezhi and Schuurmans, Dale and Bosma, Maarten and Ichter, Brian and Xia, Fei and Chi, Ed and Le, Quoc and Zhou, Denny",
  booktitle = "Conference on Neural Information Processing Systems (NeurIPS)",
  year = 2022,
  month = dec,
  pages = "24824--24837",
  url = "https://proceedings.neurips.cc/paper_files/paper/2022/hash/9d5609613524ecf4f15af0f7b31abca4-Abstract-Conference.html" }

@InProceedings{yao_et_al_2023_react,
  title = "{ReAct}: Synergizing Reasoning and Acting in Language Models",
  author = "Yao, Shunyu and Zhao, Jeffrey and Yu, Dian and Du, Nan and Shafran, Izhak and Narasimhan, Karthik R and Cao, Yuan",
  booktitle = "International Conference on Learning Representations (ICLR)",
  year = 2023,
  month = may,
  url = "https://openreview.net/forum?id=WE_vluYUL-X" }

@InProceedings{yao_et_al_2023_tot,
  title = "Tree of Thoughts: Deliberate Problem Solving with Large Language Models",
  author = "Yao, Shunyu and Yu, Dian and Zhao, Jeffrey and Shafran, Izhak and Griffiths, Tom and Cao, Yuan and Narasimhan, Karthik",
  booktitle = "Conference on Neural Information Processing Systems (NeurIPS)",
  year = 2023,
  month = dec,
  url = "https://proceedings.neurips.cc/paper_files/paper/2023/hash/271db9922b8d1f4dd7aaef84ed5ac703-Abstract-Conference.html" }

@Article{yuksekgonul_et_al_2025,
  title = "Optimizing generative {AI} by backpropagating language model feedback",
  author = "Yuksekgonul, Mert and Bianchi, Federico and Boen, Joseph and Liu, Sheng and Lu, Pan and Huang, Zhi and Guestrin, Carlos and Zou, James",
  journal = "Nature",
  year = 2025,
  month = mar,
  volume = 639,
  number = 8055,
  pages = "609--616" }

@InProceedings{zheng_et_al_2023,
  title = "Judging {LLM}-as-a-Judge with {MT}-Bench and Chatbot Arena",
  author = "Zheng, Lianmin and Chiang, Wei-Lin and Sheng, Ying and Zhuang, Siyuan and Wu, Zhanghao and Zhuang, Yonghao and Lin, Zi and Li, Zhuohan and Li, Dacheng and Xing, Eric and Zhang, Hao and Gonzalez, Joseph E and Stoica, Ion",
  booktitle = "Conference on Neural Information Processing Systems (NeurIPS)",
  year = 2023,
  pages = "46595--46623",
  url = "https://proceedings.neurips.cc/paper_files/paper/2023/file/91f18a1287b398d378ef22505bf41832-Paper-Datasets_and_Benchmarks.pdf" }

@InProceedings{zheng_et_al_2024,
  title = "{SGLang}: Efficient Execution of Structured Language Model Programs",
  author = "Zheng, Lianmin and Yin, Liangsheng and Xie, Zhiqiang and Sun, Chuyue and Huang, Jeff and Yu, Cody Hao and Cao, Shiyi and Kozyrakis, Christos and Stoica, Ion and Gonzalez, Joseph E. and Barrett, Clark and Sheng, Ying",
  booktitle = "Conference on Neural Information Processing Systems (NeurIPS)",
  year = 2024,
  pages = "62557--62583",
  url = "https://proceedings.neurips.cc/paper_files/paper/2024/file/724be4472168f31ba1c9ac630f15dec8-Paper-Conference.pdf" }

@InProceedings{zhou_et_al_2024,
  title = "Language Agent Tree Search Unifies Reasoning, Acting, and Planning in Language Models",
  author = "Zhou, Andy and Yan, Kai and Shlapentokh-Rothman, Michal and Wang, Haohan and Wang, Yu-Xiong",
  booktitle = "International Conference on Machine Learning (ICML)",
  year = 2024,
  month = jul,
  pages = "62138--62160",
  url = "https://proceedings.mlr.press/v235/zhou24r.html" }

@article{doucet-smc-2006,
  title={Sequential {Monte} {Carlo} samplers},
  author={Del Moral, Pierre and Doucet, Arnaud and Jasra, Ajay},
  journal={J. Royal Statistical Society: Series B (Statistical Methodology)},
  volume={68},
  number={3},
  pages={411--436},
  year={2006},
  publisher={Wiley Online Library}
}

@misc{rocq-prover,
  author = {{Rocq Prover Team}},
  title = {The {Rocq} Prover},
  howpublished = {\url{https://rocq-prover.org/}},
  note = {Accessed: 2026-01-26}
}

@article{minif2f,
  title   = {MiniF2F: a cross-system benchmark for formal Olympiad-level mathematics},
  author  = {Zheng, Kunhao and Han, Jesse Michael and Polu, Stanislas},
  journal = {arXiv preprint arXiv:2109.00110},
  year    = {2021}
}

@inproceedings{dsp,
  title     = {Draft, Sketch, and Prove: Guiding Formal Theorem Provers with Informal Proofs},
  author    = {Albert Q. Jiang and Sean Welleck and Jin Peng Zhou and Wenda Li and Jiacheng Liu and Mateja Jamnik and Timothée Lacroix and Yuhuai Wu and Guillaume Lample},
  booktitle = "International Conference on Learning Representations (ICLR)",
  year      = {2022},
  url = "https://openreview.net/forum?id=SMa9EAovKMC" }

@article{atp,
  title   = {Generative language modeling for automated theorem proving},
  author  = {Polu, Stanislas and Sutskever, Ilya},
  journal = {arXiv preprint arXiv:2009.03393},
  year    = {2020}
}

@inproceedings{copra,
  title     = {An in-context learning agent for formal theorem-proving},
  author    = {Thakur, Amitayush and Tsoukalas, George and Wen, Yeming and Xin, Jimmy and Chaudhuri, Swarat},
  booktitle = "Conference on Language Modeling (COLM)",
  year = 2024,
  month = oct }

@article{magnushammer,
  title   = {Magnushammer: A transformer-based approach to premise selection},
  author  = {Miku{\l}a, Maciej and Tworkowski, Szymon and Antoniak, Szymon and Piotrowski, Bartosz and Jiang, Albert Qiaochu and Zhou, Jin Peng and Szegedy, Christian and Kuci{\'n}ski, {\L}ukasz and Mi{\l}o{\'s}, Piotr and Wu, Yuhuai},
  journal = {arXiv preprint arXiv:2303.04488},
  year    = {2023}
}

@article{poetry,
  title   = {Proving Theorems Recursively},
  author  = {Wang, Haiming and Xin, Huajian and Liu, Zhengying and Li, Wenda and Huang, Yinya and Lu, Jianqiao and Yang, Zhicheng and Tang, Jing and Yin, Jian and Li, Zhenguo and others},
  journal = {arXiv preprint arXiv:2405.14414},
  year    = {2024}
}

@article{hilbert,
  title   = {Hilbert: Recursively Building Formal Proofs with Informal Reasoning},
  author  = {Varambally, Sumanth and Voice, Thomas and Sun, Yanchao and Chen, Zhifeng and Yu, Rose and Ye, Ke},
  journal = {arXiv preprint arXiv:2509.22819},
  year    = {2025}
}

@article{minif2f-rocq,
  author    = {Jules Viennot and Guillaume Baudart and Emilio Jes{\'{u}}s Gallego Arias and Marc Lelarge},
  title     = {MiniF2F in Rocq: Automatic Translation Between Proof Assistants - A Case Study},
  journal   = {CoRR},
  volume    = {abs/2503.04763},
  year      = {2025}
}

@InProceedings{hendrycksmath2021,
  title={Measuring Mathematical Problem Solving With the MATH Dataset},
  author={Dan Hendrycks and Collin Burns and Saurav Kadavath and Akul Arora and Steven Basart and Eric Tang and Dawn Song and Jacob Steinhardt},
  booktitle = "NeurIPS Datasets and Benchmarks Track",
  year = 2021,
  month = dec,
  url = "https://openreview.net/forum?id=7Bywt2mQsCe" }

@misc{math-500,
  author = {{Math-500 Dataset}},
  title = {Math-500 Benchmark Leaderboard},
  howpublished = {\url{https://artificialanalysis.ai/evaluations/math-500}},
}

@InProceedings{jain2024livecodebenchholisticcontaminationfree,
  title = "{LiveCodeBench}: Holistic and Contamination Free Evaluation of Large Language Models for Code",
  author = "Jain, Naman and Han, King and Gu, Alex and Li, Wen-Ding and Yan, Fanjia and Zhang, Tianjun and Wang, Sida and Solar-Lezama, Armando and Sen, Koushik and Stoica, Ion",
  booktitle = "International Conference on Learning Representations (ICLR)",
  year = 2025,
  month = may,
  url = "https://openreview.net/forum?id=chfJJYC3iL" }

\clearpage
\appendix

\section{Example Walk-Through}
\label{sec:walkthrough}
This section illustrates the execution of the PPDL program from \Cref{ppdl} using two different inference algorithms: Importance Sampling (IS) and Sequential Monte Carlo (SMC). Both examples use 5 particles to solve the centered hexagonal number problem, where the task is to generate a Python function that computes the $n$th centered hexagonal number using the formula $1 + 3n(n-1)$.

\paragraph{Importance Sampling execution (\Cref{exec-is}).}
In IS, all particles execute independently from start to finish. Each particle generates a plan (first row), which is then scored using an LLM-as-a-judge constraint (second row). The particles then generate code solutions (third row), which are scored using a linter-based constraint (fourth row).

The key insight from this execution is how the scoring mechanism differentiates solution quality. Particles~0 and~4 generate correct solutions with the formula \lstinline{1 + 3*n*(n-1)}, and receive very low penalty scores (approximately $-0.69$ after both scoring steps). Particles~1, 2, and~3 generate incorrect formulas and receive much higher penalty scores (ranging from $-14.25$ to $-17.44$).

When these scores are normalized into probabilities, particles~0 and~4 each have approximately 50\% probability, while the incorrect solutions have negligible probabilities (less than $0.0004\%$ combined). This demonstrates how IS uses scoring to amplify the probability of correct solutions. Without scoring (i.e., using majority voting), the probability of sampling a correct solution would be only 0.4, since only 2 out of 5 particles found the correct formula.

\paragraph{Sequential Monte Carlo execution (\Cref{exec-smc}).}
SMC differs from IS by introducing a resampling step after the first scoring point. Initially, all 5 particles generate plans and receive scores (first two rows). Particles~1 and~2 have significantly better scores ($-0.000\,000\,002$ and $-0.000\,000\,009$) because their plans contain the correct formula, while the other particles have much worse scores.

The resampling step (indicated by the arrows in \Cref{exec-smc}, similar to \Cref{fig:commuting} from the introduction) duplicates the promising particles~1 and~2 while discarding the poorly-scoring particles~0, 3, and~4. After resampling, the particle population consists of three copies derived from particle~1 and two copies derived from particle~2. These resampled particles then continue execution to generate code solutions~(third row).

Crucially, all five particles in the second phase now generate correct solutions, because they all descended from particles with good plans. After the final scoring step, all particles have identical scores ($-0.69$), resulting in a uniform distribution where each solution has equal probability. This execution demonstrates SMC's ability to focus computational resources on promising execution paths early in the computation, leading to higher-quality final results.

The contrast between these two executions highlights the fundamental difference between IS and SMC: IS evaluates all particles independently and uses scoring only at the end to weigh results, while SMC actively steers the particle population toward promising regions of the solution space through intermediate resampling steps.

\begin{table*}[ht]
\caption{\label{exec-is}Example of execution of the program provided in \Cref{ppdl} using Importance Sampling. The columns represent the state of the different particles, and the rows represent the values of the variables through the execution.
The circles visualize probability magnitudes as in \Cref{fig:commuting}.
There is a probability of 0.999\,999\,453 to sample a correct solution (particles 0 and 4) in this distribution.
If we ignore the scores, this trace corresponds to a majority voting execution. In this case, the probability of sampling a correct solution is 0.4.}
\centering
\renewcommand{\arraystretch}{1.2}
\begin{tabularx}{\textwidth}{|l|X|X|X|X|X|}
\hline
\textbf{Variable} &
\textbf{Particle 0} &
\textbf{Particle 1} &
\textbf{Particle 2} &
\textbf{Particle 3} &
\textbf{Particle 4} \\
\hline

{\ttfamily\small plan} &
\dots~
- The formula `3*n*(n+1) + 1` is a more direct representation of the nth centered hexagonal number.
~\dots &
\dots~
The formula to calculate the nth centered hexagonal number is given by: $1 + 6 * (1 + 2 + ... + n)$, which simplifies to $1 + 6 * (n * (n + 1) / 2)$.
~\dots &
\dots~
The formula for the nth centered hexagonal number is given by: 1 + 6 * (1 + 2 + ... + n), which simplifies to 1 + 6 * (n * (n + 1) / 2).
~\dots &
\dots~
The formula for the nth centered hexagonal number is given by: 1 + 6 * (1 + 2 + ... + n), which simplifies to 1 + 6 * (n * (n + 1) / 2).
~\dots &
\dots~
- The correct formula for the nth centered hexagonal number is `1 + 3*n*(n-1)`.
~\dots
\\
\hline

{\ttfamily\small score} &
$-0.000\,000\,001$ &
$-14.750\,000\,035$ &
$-16.750\,000\,053$ &
$-14.250\,001\,005$ &
$-0.000\,000\,012$ \\
(prob) &
$0.499\,999\,729$ \probcircle{probhig} &
$0.000\,000\,196$ \probcircle{problow} &
$0.000\,000\,027$ \probcircle{problow} &
$0.000\,000\,324$ \probcircle{problow} &
$0.499\,999\,724$ \probcircle{probhig}\\
\hline

{\ttfamily\small solution} &
\begin{minipage}[t]{\linewidth}\ttfamily\scriptsize
def centered\_hex...\\
    """\\
    ...\\
    """\\
    return 3*n*(n-1) + 1
\end{minipage}
&
\begin{minipage}[t]{\linewidth}\ttfamily\scriptsize
def centered\_hex...\\
    """\\
    ...\\
    """\\
    if not isinstance(n, int):\\
        raise TypeError("Input must be an integer.")\\
    if n < 0:\\
        raise ValueError("Input must be a non-negative integer.")\\
    return 1 + 3 * n * (n + 1)\\
\end{minipage}
&
\begin{minipage}[t]{\linewidth}\ttfamily\scriptsize
def centered\_hex...\\
    """\\
    ...\\
    """\\
    \# Calculate the nth centered hexagonal number using the formula\\
    return 1 + 6 * (n * (n + 1) // 2)\\
\end{minipage}
&
\begin{minipage}[t]{\linewidth}\ttfamily\scriptsize
def centered\_hex...\\
    """\\
    ...\\
    """\\
    \# Calculate the sum of the first n natural numbers\\
    sum\_of\_n = n * (n + 1) // 2\\
    \\
    \# Calculate the nth centered hexagonal number\\
    result = 1 + 6 * sum\_of\_n\\
    \\
    return result\\
\end{minipage}
&
\begin{minipage}[t]{\linewidth}\ttfamily\scriptsize
def centered\_hex...\\
    return 1 + 3*n*(n-1)
\end{minipage}
\\
\hline

{\ttfamily\small score} &
$-0.693\,147\,182$ &
$-15.443\,147\,216$ &
$-17.443\,147\,234$ &
$-14.943\,148\,186$ &
$-0.693\,147\,192$ \\
(prob) &
$0.499\,999\,729$ \probcircle{probhig} &
$0.000\,000\,196$ \probcircle{problow} &
$0.000\,000\,027$ \probcircle{problow} &
$0.000\,000\,324$ \probcircle{problow} &
$0.499\,999\,724$ \probcircle{probhig} \\
\hline
\end{tabularx}
\end{table*}

\begin{table*}[ht]
\caption{\label{exec-smc}Example of execution of the program provided in \Cref{ppdl} using SMC. The arrows represents the resampling step. In this execution, all the particles compute a correct solution and they all have the same probability. After the the first \lstinline{factor}, the resampling step duplicates particles~1 and~2 who have a plan that contains the correct formula.}
\centering
\renewcommand{\arraystretch}{1.2}
\begin{tabularx}{\textwidth}{|l|X|X|X|X|X|}
\hline
\textbf{Variable} &
\textbf{Particle 0} &
\textbf{Particle 1} &
\textbf{Particle 2} &
\textbf{Particle 3} &
\textbf{Particle 4} \\
\hline

{\ttfamily\small plan} &
\dots~
Simplifying this gives 1 + 3n(n + 1) = 3n\^{}2 + 3n + 1.
~\dots &
\dots~
By substituting the sum of the first (n-1) natural numbers into the expression, we get `1 + 6*((n-1)*n/2)`, which simplifies to `1 + 3*n*(n-1)`.
~\dots &
\dots~
The formula to calculate the nth centered hexagonal number is given by: $1 + 6 + 12 + ... + 6(n-1)$, which simplifies to $3n^2 - 3n + 1$.
~\dots &
\dots~
Substituting the sum of the first `n` natural numbers into the formula for the nth centered hexagonal number gives `1 + 6 * (n * (n + 1) / 2)`.
~\dots &
\dots~
The formula can be simplified to 1 + 6(n(n+1)/2).
~\dots
\\
\hline

{\ttfamily\small score} &
\makebox[\linewidth][c]{$-10.75$} &
\makebox[\linewidth][c]{$-0.000\,000\,002$} &
\makebox[\linewidth][c]{$-0.000\,000\,009$} &
\makebox[\linewidth][c]{$-15.99$} &
\makebox[\linewidth][c]{$-18.75$} \\
(prob) &
\phead{p0}{$0.000\,010\,723$ \probcircle{problow}} &
\phead{p1}{$0.499\,994\,610$ \probcircle{probhig}} &
\phead{p2}{$0.499\,994\,607$ \probcircle{probhig}} &
\phead{p3}{$0.000\,000\,057$ \probcircle{problow}} &
\phead{p4}{$0.000\,000\,004$ \probcircle{problow}} \\
\hline

\noalign{\vskip 5.0em}

\hline
{\ttfamily\small solution} &
\emptytarget{q1}\vspace{-1.2em}
\begin{minipage}[t]{\linewidth}\ttfamily\scriptsize
def centered\_hex...\\
    """\\
    ...\\
    """\\
    if not isinstance(n, int) or n <= 0:\\
        raise ValueError("n must be a positive integer")\\
    return 1 + 3*n*(n-1)\\
\end{minipage}
&
\emptytarget{q2}\vspace{-1.2em}
\begin{minipage}[t]{\linewidth}\ttfamily\scriptsize
def centered\_hex...\\
    return 1 + 3*n*(n-1)
\end{minipage}
&
\emptytarget{q3}\vspace{-1.2em}
\begin{minipage}[t]{\linewidth}\ttfamily\scriptsize
def centered\_hex...\\
    """\\
    ...\\
    """\\
    \# Calculate the nth centered hexagonal number using the formula: 3n\^{}2 - 3n + 1\\
    return 3*n**2 - 3*n + 1\\
\end{minipage}
&
\emptytarget{q4}\vspace{-1.2em}
\begin{minipage}[t]{\linewidth}\ttfamily\scriptsize
def centered\_hex...\\
    return 1 + 3*n*(n-1)
\end{minipage}
&
\emptytarget{q5}\vspace{-1.2em}
\begin{minipage}[t]{\linewidth}\ttfamily\scriptsize
def centered\_hex...\\
    """\\
    ...\\
    """\\
    return 3*n**2 - 3*n + 1\\
\end{minipage}
\\
\hline

{\ttfamily\small score} &
\makebox[\linewidth][c]{$-0.69$} &
\makebox[\linewidth][c]{$-0.69$} &
\makebox[\linewidth][c]{$-0.69$} &
\makebox[\linewidth][c]{$-0.69$} &
\makebox[\linewidth][c]{$-0.69$} \\
(prob) &
\makebox[\linewidth][c]{0.2 \probcircle{probmed}} &
\makebox[\linewidth][c]{0.2 \probcircle{probmed}} &
\makebox[\linewidth][c]{0.2 \probcircle{probmed}} &
\makebox[\linewidth][c]{0.2 \probcircle{probmed}} &
\makebox[\linewidth][c]{0.2 \probcircle{probmed}} \\
\hline
\end{tabularx}

\begin{tikzpicture}[
  remember picture,
  overlay,
  every path/.style={
-{Stealth},
    thick,
    shorten >=2pt,
    shorten <=2pt
  }
]

\draw[shorten >=4pt,shorten <=4pt] (p1.south) -- (q1.north);
\draw                              (p1.south) -- (q2.north);
\draw                              (p2.south) -- (q3.north);
\draw[shorten >=6pt,shorten <=8pt] (p1.south) -- (q4.north);
\draw[shorten >=6pt,shorten <=8pt] (p2.south) -- (q5.north);
\end{tikzpicture}
\end{table*}
 \section{Possible alternatives for \Cref{base,ppdl}}
\label{sec:alternative}

\begin{figure}[t]
\begin{minipage}[b]{0.8\textwidth}
{
\begin{lstlisting}[language=python]
llm = "watsonx/meta-llama/llama-4-maverick-..."
problem_statement = (
 "Write a function to find nth centered hexagonal number.\n"
 "assert centered_hexagonal_number(10) == 271\n"
)

messages = [
 {
  "content": (
   "Generate an English plan for how to generate code "
   f"for the following problem: { problem_statement }"
  ),
  "role": "user",
 }
]
response = completion(model=llm, messages=messages)
plan = response.choices[0].message
messages = messages + [ plan ]
messages = messages + [
 {
  "content": (
   "Generate a complete executable Python function "
   "definition corresponding to the above plan and "
   "problem. Generate only a single function definition."
  ),
  "role": "user",
 }
]
response = completion(model=llm, messages=messages)
solution_str = response.choices[0].message["content"]
solution_match = re.fullmatch(
  r"(.|\n)*```python\n(?P<code>(.|\n)*?)```(.|\n)*",
  solution_str, flags=re.M
)
solution = solution_match.group("code")
\end{lstlisting}}
\end{minipage}
\caption{\label{base-litellm} Implementation in Python using LiteLLM of \Cref{base}.}
\end{figure}

\begin{figure}[t]
\begin{minipage}[b]{0.8\textwidth}
{
\begin{lstlisting}[language=python]
llm = "watsonx/meta-llama/llama-4-maverick-..."
problem_statement = (
 "Write a function to find nth centered hexagonal number.\n"
 "assert centered_hexagonal_number(10) == 271\n"
)

solutions_and_scores = []
for _ in range(n):
  score = 0
  messages = [
   {
    "content": (
     "Generate an English plan for how to generate code "
     f"for the following problem: { problem_statement }"
    ),
    "role": "user",
   }
  ]
  response = completion(model=llm, messages=messages)
  plan = response.choices[0].message
  messages = messages + [ plan ]
  constraint = (
      "This plan for the following problem is correct.\n"
      f"{problem_statement}\n"
  )
  score = score + utils.llm_judge(llm, plan, constraint)
  messages = messages + [
   {
    "content": (
     "Generate a complete executable Python function "
     "definition corresponding to the above plan and "
     "problem. Generate only a single function definition."
    ),
    "role": "user",
   }
  ]
  response = completion(model=llm, messages=messages)
  solution_str = response.choices[0].message["content"]
  solution_match = re.fullmatch(
    r"(.|\n)*```python\n(?P<code>(.|\n)*?)```(.|\n)*",
    solution_str, flags=re.M
  )
  solution = solution_match.group("code")
  score = score + utils.score_errors_and_warnings(solution)
  solutions_and_scores.append((solution, score))

dist = Categorical(solutions_and_scores)
\end{lstlisting}}
\end{minipage}
\caption{\label{litellm-is} Implementation in Python using LiteLLM of \Cref{ppdl} with IS probabilistic inference.}
\end{figure}

\begin{figure}[t]
\begin{minipage}[b]{0.8\textwidth}
{
\begin{lstlisting}[language=python]
llm = "watsonx/meta-llama/llama-4-maverick-..."
problem_statement = (
 "Write a function to find nth centered hexagonal number.\n"
 "assert centered_hexagonal_number(10) == 271\n"
)

states = []
scores = []
for _ in range(n):
  score = 0
  messages = [
   {
    "content": (
     "Generate an English plan for how to generate code "
     f"for the following problem: { problem_statement }"
    ),
    "role": "user",
   }
  ]
  response = completion(model=llm, messages=messages)
  plan = response.choices[0].message
  messages = messages + [ plan ]
  constraint = (
      "This plan for the following problem is correct.\n"
      f"{problem_statement}\n"
  )
  score = score + utils.llm_judge(llm, plan, constraint)
  states.append(messages)
  scores.append(score)

states = resample(states, scores)

solutions_and_scores = []
for messages in states:
  score = 0
  messages = messages + [
   {
    "content": (
     "Generate a complete executable Python function "
     "definition corresponding to the above plan and "
     "problem. Generate only a single function definition."
    ),
    "role": "user",
   }
  ]
  response = completion(model=llm, messages=messages)
  solution_str = response.choices[0].message["content"]
  solution_match = re.fullmatch(
    r"(.|\n)*```python\n(?P<code>(.|\n)*?)```(.|\n)*",
    solution_str, flags=re.M
  )
  solution = solution_match.group("code")
  score = score + utils.score_errors_and_warnings(solution)
  solutions_and_scores.append((solution, score))

dist = Categorical(solutions_and_scores)
\end{lstlisting}}
\end{minipage}
\caption{\label{litellm-smc} Implementation in Python using LiteLLM of \Cref{ppdl} with SMC probabilistic inference.}
\end{figure}

\begin{figure}[t]
\begin{minipage}[b]{0.8\textwidth}
{
\begin{lstlisting}[language=python]
llm = WatsonxLLM(
    model_id="meta-llama/llama-4-maverick-...",
    ...
)
parser = StrOutputParser()

problem_statement = (
 "Write a function to find nth centered hexagonal number.\n"
 "assert centered_hexagonal_number(10) == 271\n"
)

plan_prompt = HumanMessage(
  content=(
   "Generate an English plan for how to generate code "
   f"for the following problem: {problem_statement}"
  )
)

plan_inputs = [[plan_prompt]] * n
plan_responses = llm.batch(plan_inputs)

states = []
scores = []
for response in plan_responses:
  messages = [plan_prompt, response]
  plan_text = parser.invoke(response)
  constraint = (
   "This plan for the following problem is correct.\n"
   f"{problem_statement}\n"
  )
  score = utils.llm_judge(llm, plan_text, constraint)
  states.append(messages)
  scores.append(score)

states = resample(states, scores)

code_prompt = HumanMessage(
 content=(
  "Generate a complete executable Python function "
  "definition corresponding to the above plan and "
  "problem. Generate only a single function definition."
 )
)

code_inputs = [
  messages + [code_prompt] for messages in states
]
code_responses = llm.batch(code_inputs)

solutions_and_scores = []
for response in code_responses:
  solution_str = parser.invoke(response)
  solution_match = re.fullmatch(
    r"(.|\n)*```python\n(?P<code>(.|\n)*?)```(.|\n)*",
    solution_str, flags=re.M,
  )
  solution = solution_match.group("code")
  score = utils.score_errors_and_warnings(solution)
  solutions_and_scores.append((solution, score))

dist = Categorical(solutions_and_scores)
\end{lstlisting}}
\end{minipage}
\caption{\label{langchain-smc} Implementation in Python using LangChain of \Cref{ppdl} with SMC probabilistic inference.}
\end{figure}

This section demonstrates how the PPDL programs from \Cref{base,ppdl} can be implemented using standard Python libraries. These implementations illustrate both the expressiveness advantages of PPDL and its orthogonality between program and inference scaling startegy. We present four implementations: a baseline using LiteLLM, importance sampling (IS) and sequential Monte Carlo (SMC) variants with LiteLLM, and an SMC implementation using LangChain that exploits parallelism.

First, \Cref{base-litellm} corresponds to the implementation of \Cref{base} using LiteLLM (\url{https://github.com/BerriAI/litellm}) to handle LLM calls. This code has the same structure as the PDL code but requires explicit message building and accumulation~(Lines~7-15, and 18-28). It also requires accessing the LLM response data structure~(Lines~17 and~30).

\Cref{litellm-is} presents an implementation of \Cref{ppdl} using the IS inference algorithm. The implementation wraps the baseline code in a loop~(Line~8) that creates $n$ independent particles. Each particle maintains a log-probability score~(Lines~26 and~44) that is used to construct a weighted categorical distribution over solutions~(Line~47).  This implementation is purely sequential, in particular, each model call is done one after the other.

To implement \Cref{ppdl} with SMC, the IS code is restructured as shown in \Cref{litellm-smc}. The key difference is that particle execution is split into two phases separated by a resampling step~(Line~31): the first loop~(Lines~9-29) generates and scores plans,  the second loop~(Lines~34-54) generates code. This allows SMC to focus computational resources on promising particles after the first scoring point. The \lstinline[language=python]{resample} function samples $n$ states from the distribution of particle states.

\begin{lstlisting}[language=python,basicstyle=\footnotesize\ttfamily,]
def resample(states, scores):
  n = len(states)
  d = Categorical(zip(states, scores))
  return [d.sample() for _ in range(n)]
\end{lstlisting}

The SMC algorithm can be implemented using prompt programming frameworks like LangChain, as shown in \Cref{langchain-smc}. This implementation exploits LangChain's batch API~(Lines~21 and 49) to execute particles in parallel, significantly reducing wall-clock time. The batch API sends multiple prompts to the LLM simultaneously.

Compared to these implementations, PPDL provides a high-level abstraction that separates the probabilistic specification from inference implementation details, allowing the runtime to automatically optimize execution (e.g., through parallelism) without programmer intervention.
 \section{Semantics}\label{sec:semantics}

In this section, we present semantics of PPDL.

The syntax of PPDL using the flow-style of YAML is the following:

{\small
$
\begin{array}{rc@{~}l}
    \mit{pdl} & ::= &\phantom{\mid}
      \begin{array}[t]{@{}l@{~}l@{~}r@{}}
      \mtt{\{} 
        & \mtt{defs:} \jsobj{x\mtt{:}\mit{pdl}\mtt{,} \dots\mtt{,} x\mtt{:}\mit{pdl}} \mtt{,}
      \\& \mit{body}\mtt{,}
      \\& \mtt{parser:} \mit{parser}\mtt{,}
      \\& \mtt{contribute:} \mit{contribute}\mtt{,}
      \\& \mtt{def:} \mit{x}
      & \mtt{\}}
      \end{array}
    \end{array}
    $
}

{\small
$$
\begin{array}{rc@{~}l}
\mit{body} & ::= &\phantom{\mid}
    \pdldata{\mit{expr}}
    \\ &&\mid
    \pdlcode{\mit{pdl}}{\mit{string}}
    \\ &&\mid
    \pdlmodel{\mit{expr}}{\mit{pdl}}
    \\ &&\mid
    \pdlif{\mit{expr}}{\mit{pdl}}{\mit{pdl}}
    \\ &&\mid
    \pdlsequence{\jsarr{\mit{pdl}\mtt{,} \dots\mtt{,} \mit{pdl}}}{\mit{join}}
    \\ &&\mid
    \pdlwhile{\mit{expr}}{\mit{pdl}}{\mit{join}}
    \\ &&\mid
    \pdlfunction{\mit{types}}{\mit{pdl}}
    \\ &&\mid
    \pdlcall{\mit{expr}}{\mit{expr}}
    \\ &&\mid
    \pdlfactor{\mit{expr}}
\\\\

    \mit{expr} & ::= &\phantom{\mid}
    \jsnull
    \mid
    \mit{bool}
    \mid
    \mit{number}
    \mid
    \mit{string}
    \mid
    \jinja{\mit{jinja\_expr}}
    \\ &&\mid \jsarr{\mit{expr} \mtt{,} \dots \mtt{,} \mit{expr}}
    \mid \jsobj{x\mtt{:}\mit{expr}\mtt{,} \dots\mtt{,} x\mtt{:}\mit{expr}}
\end{array}
$$}

A program is a block comprising a set of variable definitions, the block body (i.e., the instruction to evaluate), a parser to extract the result from the return value of the body, a flag indicating whether the result contributes to the context, and the name of a variable to store this result.

\clearpage

The body of a block can be an expression~(\mtt{data}), some code in Python, Shell, Jinja, or PDL, an LLM call with its input, a conditional, a sequence of blocks with a join operator to gather result values, a loop with a join operator to gather the results of each iteration, a function definition, a function call, or a factor statement to update the score.
The $\mit{join}$ operator is a function that takes a list as input and returns a value. In the concrete syntax of PPDL, 
$\jsobj{\mtt{join:} \jsobj{\mtt{as:}\mtt{text}}}$ corresponds to the string concatenation, $\jsobj{\mtt{join:} \jsobj{\mtt{as:}\mtt{lastOf}}}$ returns the last element of the list, and $\jsobj{\mtt{join:} \jsobj{\mtt{as:}\mtt{array}}}$ is the identity function.

Types~($\mit{type}$) are a subset of JSON Schema, the parsers~($\mit{parser}$) are regular expressions, and the contribute flag is defined as follows:
$$
\begin{array}{rcl}
    \mit{contribute} & ::= &\phantom{\mid}
    \jsarr{}
    \mid
    \jsarr{\pdlcontext}
\end{array}
$$

The ideal semantics of a PPDL program~$p$ is defined by the infinite number of executions of the weighted sampler associated to the program~$p$ and the normalization of the weights to obtain a categorical distribution:
{\small
\begin{mathpar}
\inferrule
{
 \left\{ \pdlsempstar{S}{p}{S_i}{v_i} \right\}_{1 \le i \le N}\\
\left\{ w_i = \exp(S_i'[\texttt{pdl\_score}]) \right\}_{1 \le i \le N}\\
 W = \textstyle\sum_{1 \le i \le N} w_i
}
{\pdlsemd{S}{p}{\lambda U. \textstyle\sum_{1 \le i \le N}(w_i / W) \times \delta_{v_i}(U)}}
\end{mathpar}
}
The reduction~$\Rightarrow^*$ is the transitive closure of the weighted sampler semantics and~${N \rightarrow \infty}$.

The semantics of weighted sampler~$p$ is defined \Cref{fig:pdlsemp} by the reduction $\pdlsemp{S}{p}{S'}{\mit{p\_or\_v}}$ such that, given an environment~$S$, the program~$p$ evaluates to the updated environment~$S'$, and rewrites into a new program or a value.

The semantics of variable definitions is defined \Cref{fig:pdlsemdefs} by the reduction $\pdlsemdefs{S}{\mit{defs}}{S'}{\mit{defs}'}$  such that, given an environment~$S$, the definitions~$\mit{defs}$ evaluates to the updated environment~$S'$, and rewrites into a new set of definitions~$\mit{defs}'$.

The semantics of a block body~$b$ is defined \Cref{fig:pdlsembb} by the reduction $\pdlsembb{S}{b}{S'}{\mit{b\_or\_v}}$ such that, given an environment~$S$, the block body~$b$ evaluates to the updated environment~$S'$, and rewrites into a new block body or a value.

The semantics of an expression~$e$ is defined \Cref{fig:pdlseme} by the reduction $\pdlseme{S}{e}{v}$ such that, given an environment~$S$, the expression~$e$ evaluates a value. The reduction ${\pdlseml{l}{S}{c}{v}}$ uses the semantics of the language~$l$ to evaluate the code~$c$ to a value~$v$. Evaluation of expressions and external code are not supposed to do side effects.

The semantics of parsing a value~$v$ by the parser~$\mit{parser}$ to extract a value~$v'$ is defined by the semantics of Python regular expressions and is noted $\pdlsemparse{\mit{parser}}{v}{v'}$.

The semantics of contribute is defined \Cref{fig:pdlsemcontrib} by the reduction $\pdlsemcontrib{S}{v}{\mit{contribute}}{S'}$ such that, given an environment~$S$, the value~$v$, and the contribute flag~$\mit{contribute}$ produces a new environment~$S'$.

\begin{figure*}
\centering
\small
\begin{mathpar}
\inferrule
{\pdlsemdefs{S}{\mit{defs}}{S_1}{\jsobj{}} \\
 \pdlsembb{S_1}{\mit{block\_body}}{S_2}{v} \\
 \pdlsemparse{\mit{parser}}{v}{v'} \\
 \pdlsemcontrib{S_2}{v'}{\mit{contribute}}{S_3} \\
 S' = S_3[x \leftarrow v']
}
{\pdlsemp{S}{
      \mtt{\{}
      \mtt{defs:} \mit{defs}\mtt{,}
      \mit{block\_body}\mtt{,}
      \mtt{parser:} \mit{parser}\mtt{,}
      \mtt{contribute:} \mit{contribute}\mtt{,}
      \mtt{def:} \mit{x}
      \mtt{\}}
}{S'}{v'}}

\inferrule
{\pdlsemdefs{S}{\mit{defs}}{S'}{\mit{defs'}} \\
 \mit{defs'} \not= \jsobj{}
}
{
\ensuremath{
\begin{array}{c}
      \mtt{\{}
      \mtt{defs:} \mit{defs}\mtt{,}
      \mit{block\_body}\mtt{,}
      \mtt{parser:} \mit{parser}\mtt{,}
      \mtt{contribute:} \mit{contribute}\mtt{,}
      \mtt{def:} \mit{x}
      \mtt{\}}
      / S
    \\ \Rightarrow \\
      \mtt{\{}
      \mtt{defs:} \mit{defs'}\mtt{,}
      \mit{block\_body}\mtt{,}
      \mtt{parser:} \mit{parser}\mtt{,}
      \mtt{contribute:} \mit{contribute}\mtt{,}
      \mtt{def:} \mit{x}
      \mtt{\}}
      / S'
\end{array}}
}

\inferrule
{\pdlsemdefs{S}{\mit{defs}}{S_1}{\jsobj{}} \\
 \pdlsembb{S_1}{\mit{block\_body}}{S'}{b} \\
}
{
\ensuremath{
\begin{array}{c}
      \mtt{\{}
      \mtt{defs:} \mit{defs}\mtt{,}
      \mit{block\_body}\mtt{,}
      \mtt{parser:} \mit{parser}\mtt{,}
      \mtt{contribute:} \mit{contribute}\mtt{,}
      \mtt{def:} \mit{x}
      \mtt{\}}
      / S
    \\ \Rightarrow \\
      \mtt{\{}
      \mtt{defs:} \jsobj{}\mtt{,}
      \mit{b}\mtt{,}
      \mtt{parser:} \mit{parser}\mtt{,}
      \mtt{contribute:} \mit{contribute}\mtt{,}
      \mtt{def:} \mit{x}
      \mtt{\}}
      / S'
\end{array}}
}
\end{mathpar}
\caption{Semantics of a block}
\label{fig:pdlsemp}
\end{figure*}

\begin{figure*}
\centering
\small
\begin{mathpar}
\inferrule
{~}
{\pdlsemdefs{S}{\jsobj{}}{S}{\jsobj{}}}

\inferrule
{\pdlsemp{S}{\mit{pdl_1}}{S_1}{\mit{pdl_1'}} \\
}
{\pdlsemdefs{S}{\jsobj{x_1\mit{:} \mit{pdl_1}\mtt{,} x_2\mit{:} \mit{pdl_2}\mtt{,} \dots\mtt{,} x_n\mit{:} \mit{pdl_n}}}{S_1}{\jsobj{x_1\mit{:} \mit{pdl_1'}\mtt{,} x_2\mit{:} \mit{pdl_2}\mtt{,} \dots\mtt{,} x_n\mit{:} \mit{pdl_n}}}}

\inferrule
{\pdlsemp{S}{\mit{pdl_1}}{S_1}{v_1} \\
 \pdlsemdefs{S_1[x_1 \leftarrow v_1]}{\jsobj{x_2\mit{:} \mit{pdl_2}\mtt{,} \dots\mtt{,} x_n\mit{:} \mit{pdl_n}}}{S'}{\mit{defs}}
}
{\pdlsemdefs{S}{\jsobj{x_1\mit{:} \mit{pdl_1}\mtt{,} x_2\mit{:} \mit{pdl_2}\mtt{,} \dots\mtt{,} x_n\mit{:} \mit{pdl_n}}}{S'}{\mit{defs}}}
\end{mathpar}
\caption{Semantics of \mtt{defs} field.}
\label{fig:pdlsemdefs}
\end{figure*}

\begin{figure*}
\centering
\small
\begin{mathpar}
\inferrule {\pdlseme{S}{e}{v}}
{\pdlsembb{S}{\pdldata{e}}{S}{v}}

\inferrule {\pdlseme{S}{p}{c} \\
 \pdlseml{l}{S}{c}{v}}
{\pdlsembb{S}{\pdlcode{p}{l}}{S}{v}}

\inferrule {\pdlseme{S}{\mit{model}}{m} \\
 \pdlseme{S}{\mit{input}}{i} \\
 \pdlllm{m}{i}{v}
}
{\pdlsembb{S}{\pdlmodel{\mit{model}}{\mit{input}}}{S}{v}}

\inferrule {v = j(vs)
}
{\pdlsembb{S}{\pdlsequence[vs]{\jsarr{}}{j}}{S}{v}}

\inferrule
{\pdlsemp{S}{p_1}{S_1}{v_1} \\
 \pdlsembb{S_1}{\pdlsequence[vs + \jsarr{v_1}]{\jsarr{\mit{p_2} \dots\mtt{,} \mit{p_n}}}{j}}{S'}{p'}
}
{\pdlsembb{S}{\pdlsequence[vs]{\jsarr{\mit{p_1}\mtt{,} \mit{p_2} \dots\mtt{,} \mit{p_n}}}{j}}{S'}{p'}}

\inferrule
{\pdlsemp{S}{p_1}{S_1}{p_1'} \\
}
{\pdlsembb{S}{\pdlsequence[vs]{\jsarr{\mit{p_1}\mtt{,} \mit{p_2} \dots\mtt{,} \mit{p_n}}}{j}}{S_1}{\pdlsequence[vs]{\jsarr{\mit{p_1'} \dots\mtt{,} \mit{p_n}}}{j}}}

\inferrule {\pdlseme{S}{c}{\mtt{true}} \\
 \pdlsemp{S}{p_1}{S'}{p_1'}
}
{\pdlsembb{S}{\pdlif{c}{p_1}{p_2}}{S'}{p_1'}}

\inferrule {\pdlseme{S}{c}{\mtt{false}} \\
 \pdlsemp{S}{p_2}{S'}{p_2'}
}
{\pdlsembb{S}{\pdlif{c}{p_1}{p_2}}{S'}{p_2'}}

\inferrule {\pdlseme{S}{e}{\mtt{false}}}
{\pdlsembb{S}{\pdlwhile[vs]{e}{p}{j}}{S}{j(vs)}}

\inferrule {\pdlseme{S}{e}{\mtt{true}}\\
 \pdlsemp{S}{p}{S'}{v}}
{\pdlsembb{S}{\pdlwhile[vs]{e}{p}{j}}{S'}{\pdlwhile[vs + \jsarr{v}]{e}{p}{j}}}

\inferrule {\pdlseme{S}{e}{w}\\
 S' = S[\pdlscore \leftarrow S[\pdlscore] + w]
}
{\pdlsembb{S}{\pdlfactor{e}}{S'}{\pdldata{\jsstr{}}}}

\inferrule {~}
{\pdlsembb{S}{\pdlfunction{t}{p}}{S}{\pdlclosure{t}{p}{S}}}

\inferrule {
 \pdlseme{S}{f}{\pdlclosure{t}{\mit{p}}{S_f}}\\
 \pdlseme{S}{\mit{args}}{\mit{args}'}\\
 args' \in t\\\\
 \pdlsemp{S_f[\pdlcontext \leftarrow S[\pdlcontext], \pdlscore \leftarrow S[\pdlscore]] + \mit{args}'}{p}{S_f'}{v}\\
 S' = S[\pdlcontext \leftarrow S_f'[\pdlcontext], \pdlscore \leftarrow S_f'[\pdlscore]]
}
{\pdlsembb{S}{\pdlcall{f}{\mit{args}}}{S'}{v}}

\inferrule {
 \pdlseme{S}{f}{\pdlclosure{t}{\mit{p}}{S_f}}\\
 \pdlseme{S}{\mit{args}}{\mit{args}'}\\
 args' \in t\\\\
 \pdlsemp{S_f[\pdlcontext \leftarrow S[\pdlcontext], \pdlscore \leftarrow S[\pdlscore]] + \mit{args}'}{p}{S_f'}{p'}\\
 S' = S[\pdlcontext \leftarrow S_f'[\pdlcontext], \pdlscore \leftarrow S_f'[\pdlscore]]
}
{\pdlsembb{S}{\pdlcall{f}{\mit{args}}}{S'}{\pdlcall{\pdlclosure{t}{\mit{p'}}{S_f'}}{\mit{args}'}}}

\end{mathpar}
\caption{Semantics of block bodies.}
\label{fig:pdlsembb}
\end{figure*}

\begin{figure*}
\centering
\small
\begin{mathpar}
\inferrule
{c \in (\jsnull \cup \mit{bool} \cup \mit{{number} \mit{string}})}
{\pdlseme{S}{c}{c}}

\inferrule
{\pdlseml{\mit{jinja}}{S}{e}{v}}
{\pdlseme{S}{\jinja{e}}{v}}

\inferrule
{\left\{ \pdlseme{S}{e_i}{v_i} \right\}_{1 \le i \le N}}
{\pdlseme{S}{\jsarr{e_1\mit{,}\dots\mit{,}e_n}}{\jsarr{v_1\mit{,}\dots\mit{,}v_n}}}

\inferrule
{\left\{ \pdlseme{S}{e_i}{v_i} \right\}_{1 \le i \le N}}
{\pdlseme{S}{\jsobj{x_1\mit{:}e_1\mit{,}\dots\mit{,}x_n\mit{:}e_n}}{\jsobj{x_1\mit{:}v_1\mit{,}\dots\mit{,}{x_n}\mit{:}v_n}}}
\end{mathpar}
\caption{Semantics of expressions.}
\label{fig:pdlseme}
\end{figure*}

\begin{figure*}
\centering
\small
\begin{mathpar}
\inferrule
{~}
{\pdlsemcontrib{S}{v}{\jsarr{}}{S}}

\inferrule
{
 \mit{ctx} = S[\pdlcontext] + \jsarr{\jsobj{\mtt{content:} v}}\\
 S' = S[\pdlcontext \leftarrow \mit{ctx}]
}
{\pdlsemcontrib{S}{v}{\jsarr{\pdlcontext}}{S'}}
\end{mathpar}
\caption{Semantics of \mtt{contribute} field.}
\label{fig:pdlsemcontrib}
\end{figure*}
 
\section{Models used in experiments}\label{sec:models}

Table~\ref{tab:models} shows the selection of LLMs used in experiments.

\begin{table}[h!]
\caption{\label{tab:models}Large language models used in experiments.}
  \centerline{\small\begin{tabular}{@{}l@{~}rrrr@{}}
    \toprule
    Model          & \multicolumn{2}{c}{Size} & \multicolumn{2}{c}{Dates}\\
    \cmidrule(lr){2-3}\cmidrule(lr){4-5}
                     & total & active &  cutoff & release\\
    \midrule         
    granite4-small  &   32b &     9b & 11/2024 & 10/2025\\
    granite4-micro  &    3b &     3b &  4/2024 & 10/2025\\
    gpt-oss-120b    &  117b &     5b &  6/2024 &  8/2025\\
    gpt-oss-20b     &  21b  &     4b &  6/2024 &  8/2025\\
    llama4-maverick &  400b &    17b &  8/2024 &  4/2025\\
    llama4-scout    &  109b &    17b &  8/2024 &  4/2025\\
    \bottomrule
  \end{tabular}}
\end{table}

\section{Benchmark Code}
\label{sec:benchmark-code}

This section presents the complete PPDL programs used in the evaluation (\Cref{sec:evaluation}). Each program demonstrates how PPDL enables declarative specification of LLM workflows with probabilistic inference through \lstinline{factor} statements. The programs use LLM-as-a-judge and rule-based constraints to guide inference toward correct solutions.

\subsection{GSM8k}
\label{sec:benchmark-gsm8k}

The GSM8k program (\Cref{code:gsm8k}) solves grade-school math problems through a simple three-step workflow. First, it prompts the LLM to solve the problem and generate the final answer prefixed by \lstinline{####} (Lines~2-8). Second, it extracts the numerical answer using Python code (Lines~9-16). Third, it applies an LLM-as-a-judge \lstinline{factor} (Line~18-28) to score the solution's correctness (Lines~18-28). The \lstinline{fallback} attribute (Lines~30-31) assigns a large negative score if an error occurs, effectively filtering out malformed responses during inference.

\begin{figure}[h]
\begin{lstlisting}[language=pdl]
lastOf:
- >
  Question: ${ problem }
  Reason about this math problem and solve it. Generate 
  the final answer on the last line prefixed by `####`
- model: ${ model }
  def: solution
  parameters: ${ parameters }
- lang: python
  def: result
  code: |
    from pdl.optimize.parse_number import (
      extract_math_answer
    )
    result_line = solution.splitlines()[-1]
    result = extract_math_answer(result_line)

- defs:
    constraint: >
      Is the solution below to the following math problem
      correct?

      
      Math Problem: ${ problem }

      Solution: ${ result }
  factor:
    ${ stdlib.llm_as_judge(model, constraint, parameters) }
- ${ result }
fallback:
 factor: -100
\end{lstlisting}
\caption{\label{code:gsm8k}PPDL program for GSM8k benchmark.}
\end{figure}

\subsection{Math500}
\label{sec:benchmark-math500}

The Math500 program (\Cref{code:math500}) follows a similar structure to GSM8k but uses a more sophisticated judging strategy. After generating the solution (Lines~2-5), it uses an LLM-as-a-judge to perform line-by-line reasoning analysis (Lines~6-19), asking the judge to identify any errors in the solution's reasoning. This more detailed validation is appropriate for the harder problems in Math500, which span algebra, geometry, number theory, pre-calculus, and probability. Like GSM8k, it uses a \lstinline{fallback} to handle failures.

\begin{figure}[h]
\begin{lstlisting}[language=pdl]
lastOf:
- "Problem: ${ problem }"
- model: ${ model }
  def: solution
  parameters: ${ parameters }
- defs:
    constraint: >
      Consider the solution below to the following math
      problem. Read it line by line and reason about it
      to identify any errors in reasoning.
      If no issues are found, respond True. If any issue
      is found, respond False.
      

      Math Problem: ${ problem }

      Solution: ${ solution }
  factor:
    ${ stdlib.llm_as_judge(model, constraint, parameters) }
- ${ solution }
fallback:
 factor: -100
\end{lstlisting}
\caption{\label{code:math500}PPDL program for Math500 benchmark.}
\end{figure}

\subsection{MBPP}
\label{sec:benchmark-mbpp}

The MBPP program (\Cref{code:mbpp}) implements a multi-stage code generation workflow with three scoring points. First, it generates an English plan (Lines~5-12) and scores it using an LLM-as-a-judge (Lines~13-26). Second, it generates Python code based on the plan (Lines~27-49), with regex parsing to extract code from markdown blocks (Lines~41-43). Third, it applies multiple constraints: a rule-based check for function definitions (Line~50), a linter-based score using flake8 warnings (Line~51), and a final LLM-as-a-judge for correctness (Lines~52-66). This demonstrates PPDL's ability to combine different types of constraints—LLM-based, rule-based, and tool-based—in a single program. The program imports utility functions from \lstinline{utils.pdl} (Lines~1-3).

\begin{figure}[h]
\begin{lstlisting}[language=pdl]
defs:
  utils:
    import: utils.pdl  
lastOf:
- >
  Generate an English plan for how to generate code for
  the following problem:

  ${ problem }
- model: ${ model }
  parameters: ${ parameters }
  def: plan
- defs:
    constraint: >
      Consider the following plan for generating code for
      the coding problem below:

      Coding Problem: ${ problem }
      

      Plan: ${ plan }


      Is the plan correct?
  factor:
    ${ stdlib.llm_as_judge(model, constraint, parameters) }
- >
  Generate a complete executable Python function
  definition corresponding to the above plan for the prompt.
  Generate a single function definition only  and nothing
  else. Do not include docstrings.


  ${ problem }
- model: ${ model }
  def: response
  parameters: ${ parameters }
- defs:
    solution:
      data: ${ response }
      parser:
        regex: (.|\n)*```python\n(?P<code>(.|\n)*?)```(.|\n)*
        spec: { code: string }
- if: ${ solution == None }
  then:
    defs:
      solution: 
        data:
          code : ${ response }
- factor: ${ utils.contains_functions(solution.code) }
- factor: ${ utils.eval_number_of_warnings(solution) }
- defs:
    constraint: >
      Consider the following generated code for the coding
      problem below:

      Coding Problem: ${ problem }
      

      Code: ${ solution }
      

      Is the solution correct?

  factor:
    ${ stdlib.llm_as_judge(model, constraint, parameters) }
- ${ solution.code | default("")}

\end{lstlisting}
\caption{\label{code:mbpp}PPDL program for MBPP benchmark.}
\end{figure}

\subsection{LiveCodeBench}
\label{sec:benchmark-livecode}

The LiveCodeBench program (\Cref{code:livecode}) adapts the MBPP workflow for more challenging programming problems. It follows the same plan-then-code structure (Lines~5-12 for planning, Lines~27-42 for code generation) but requires the generated function to be named \lstinline{main} and use \lstinline{input()} for reading inputs (Lines~30-32), matching the LiveCodeBench format. The program applies three \lstinline{factor} statements: for the plan (Lines~13-26), for linter warnings (Lines~43-44), and for solution correctness (Lines~45-58).

\begin{figure}[h]
\begin{lstlisting}[language=pdl]
defs:
  utils:
    import: ../mbpp/utils.pdl
lastOf:
- >
  Generate an English plan for how to generate code for
  the following problem:

  ${ problem }
- model: ${ model }
  parameters: ${ parameters }
  def: plan
- defs:
    judge_prompt: >
      Consider the following plan for generating code for
      the coding problem below:

      Coding Problem: ${ problem }
      

      Plan: ${ plan }


      Is the plan correct?
  factor:
   ${ stdlib.llm_as_judge(model, judge_prompt, parameters) }
- >
  Generate a complete executable Python function
  definition corresponding to the above plan for the prompt.
  Generate a single function definition called `main` only
  and nothing else. Inputs to the function should be given
  using `input()`.

  Generate a single function call to the main function.
  
  ${ problem }
- model: ${ model }
  parameters: ${ parameters }
  def: solution
  parser:
    regex: (.|\n)*```python\n(?P<code>(.|\n)*?)```(.|\n)*
    spec: { code: string }
- factor:
    ${ utils.eval_number_of_warnings(response=solution) }
- defs:
    judge_prompt: >
      Consider the following generated code for the
      coding problem below:

      Coding Problem: ${ problem }
      

      Code: ${ solution }
      

      Is the solution correct?
  factor:
   ${ stdlib.llm_as_judge(model, judge_prompt, parameters) }
- ${ solution.code | default("")}
fallback:
 factor: -100
\end{lstlisting}
\caption{\label{code:livecode}PPDL program for LiveCodeBench benchmark.}
\end{figure}

\subsection{Fever}
\label{sec:benchmark-fever}

The Fever fact-checking program demonstrates PPDL's tool-use capabilities through a three-part workflow. \Cref{code:fever1} defines a Wikipedia search tool using PPDL's \lstinline{function} declaration (Lines~2-27), which wraps a Python implementation that handles disambiguation and errors. The tool signature is exposed to the LLM (Lines~28-30).

\Cref{code:fever2} implements the evidence gathering loop. After the LLM generates a response with tool calls (Lines~2-7), the program iterates over each tool call using a \lstinline{for} loop (Lines~8-58). For each search request, it validates the topic using an LLM-as-a-judge (Lines~28-45) before calling the search tool (Lines~46-48). This prevents wasting API calls on irrelevant searches. The evidence from all searches is joined with newlines (Lines~53-54).

\Cref{code:fever3} performs the final verification. It first checks whether enough evidence has been gathered using an LLM-as-a-judge (Lines~1-18), then prompts the LLM to determine if the claim is true or false based on the evidence (Lines~19-30). This three-part structure demonstrates how PPDL programs can be modularized across multiple files while maintaining a coherent probabilistic workflow.

\begin{figure}[h]
\begin{lstlisting}[language=pdl]
defs:
  search:
    description: Wikipedia search
    function:
      topic: 
        type: string
        description: Topic to be searched
    return:
      def: result
      lang: python
      code: |
        import warnings, wikipedia
        warnings.simplefilter("ignore")

        def main(topic: str, *args, **kwargs) -> str:
          try:
            return wikipedia.summary(topic)
          except wikipedia.DisambiguationError as d:
            return (
              f"\"{topic}\" may refer to one of {d.args[1]}."
              "Please retry the search with a more"
              "specific subject."
            )
          except wikipedia.WikipediaException as e:
            return str(e)

        result = main(topic)
  tools:
    data:
    - ${ search.signature }
\end{lstlisting}
\caption{\label{code:fever1}PPDL program for Fever benchmark (Part 1: tool definition).}
\end{figure}

\begin{figure}[h]
\begin{lstlisting}[language=pdl]
lastOf:
- ${ problem }
- model: ${ model }
  def: response
  modelResponse: action
  parameters: ${ dict(parameters, tools=tools)}
  parser: json
- def: evidence  
  for:
    tool_call: ${ action.choices[0].message.tool_calls }
  repeat:
    text:
    - if: ${ tool_call.function.name == "search"}
      then:
        defs: 
          args:
            data: ${ tool_call.function.arguments }
            parser: json
          topic:
            text: ${ args["topic"] }
            fallback:
              defs:
                args:
                  data: ${ args }
                  parser: json
              text: ${ args["topic"] }
        lastOf:
        - defs:
            judge_prompt: >
              Is the topic below a good topic to search
              in order to answer the following question:

              Question: ${ problem }


              Topic:

              ${ topic }


              Respond with only ('true'/'false').
          factor: >
            ${ stdlib.llm_as_judge(model,
                                   judge_prompt,
                                   parameters) }
        - call: ${ search }
          args:
            topic: ${ topic }
      else:
        factor: -100
      fallback:
        factor: -100
  join:
    with: "\n"
  fallback:
    factor: -100

\end{lstlisting}
\caption{\label{code:fever2}PPDL program for Fever benchmark (Part 2: evidence gathering).}
\end{figure}

\begin{figure}[h]
\begin{lstlisting}[language=pdl]
- defs:
    judge_prompt: >
      Is the evidence below enough to support or refute >
      the following problem:

      ${ problem }


      Evidence:

      ${ response }

      ${ evidence }


      Respond with only ('true'/'false').
  factor: 
    ${ stdlib.llm_as_judge(model, judge_prompt, parameters) }
- model: ${ model }
  parameters: ${ parameters }
  input: |
    Consider the evidence:
    ${ response }
    ${ evidence }

    Is the following statement true or false?
    ${ problem }

    Respond with only ('true'/'false').

\end{lstlisting}
\caption{\label{code:fever3}PPDL program for Fever benchmark (Part 3: final verification).}
\end{figure}

\subsection{MiniF2F}
\label{sec:benchmark-mini-f2f}

The MiniF2F theorem proving agent demonstrates PPDL's capabilities for complex iterative workflows with external tool verification. This program implements the proof/repair loop described in the case study (\Cref{sec:evaluation}): an LLM generates Rocq/Coq proofs that are verified by the Rocq proof assistant, with errors fed back to guide repair attempts.

\Cref{code:proof_agent1} shows the initialization phase. The program sets up logging infrastructure (Lines~2-7) and initializes a proof checker that interfaces with the Rocq prover using the rocq-ml-toolbox~(\url{https://github.com/LLM4Rocq/rocq-ml-toolbox}) (Lines~8-13). It then prompts the LLM with both the formal Rocq statement and an informal description of the theorem to prove (Lines~24-51), providing an example proof format to guide the LLM's output.

\Cref{code:proof_agent2} implements the core proof/repair loop using a \lstinline{repeat} block (Lines~1-55). In each iteration, the LLM generates a proof attempt (Lines~8-11), which is parsed to extract the Rocq code (Lines~12-24). The proof is then verified by calling the Rocq checker (Lines~26-28). If verification fails (Lines~30-47), a \lstinline{factor} of $-1$ is applied (Line~39) to score this attempt, and the error message is fed back to the LLM to guide the next repair attempt (Lines~40-46). This scoring mechanism allows SMC to prioritize particles with fewer errors. The loop continues until either a proof is found or the token budget is exhausted (Line~53), with a maximum iteration limit (Line~54).

\Cref{code:proof_agent3} performs final verification and result collection. After the loop terminates, it double-checks the proof correctness (Lines~7-10) and logs the final status (Lines~11-24). If the proof is complete, it records the proof, the number of steps, and the token usage (Lines~25-36); then raises an exception to signal success and terminate all parallel particles (Line~35). This early termination mechanism is crucial for efficiency: once any particle finds a valid proof, the entire computation stops, avoiding wasted token usage.

This three-part structure demonstrates PPDL's ability to express complex agent workflows with iterative refinement, external tool integration, probabilistic scoring based on intermediate feedback, and coordinated termination across parallel executions.

\lstset{
  literate={╔}{{-}}1
           {═}{{-}}1
           {╗}{{-}}1
           {║}{{|}}1
           {╚}{{-}}1
           {╝}{{-}}1
}
\begin{figure}[h]
\begin{lstlisting}[language=pdl,breaklines=true]
defs:
  logger:
    aggregator:
      file:
        ${ logdir }/logs_${name}_${pdl_particle_id}.txt
      flush: true
      mode: w
  checker: 
    lang: python
    code: |
      from proof_checker import Checker
      workspace=agent_params.get("workspace", "")
      result = Checker(workspace=workspace)
  tokens: 0
lastOf:
- text: |

    ╔════════════════════════════════════════════════╗
    ║ STARTING PROOF.                                ║
    ╚════════════════════════════════════════════════╝
    Goal: ${ formal_statement }
  contribute: [logger, stdout]
- text: |
    Please provide a proof for the following Rocq/Coq statement:

    ```Rocq

    ${ formal_statement }
    ```
    It corresponds to the following informal statement: ```${informal_statement}```

    Please use Coq/Rocq and enclose the proof in triple backticks.

    As an example, if the statement to prove is

    ```Rocq
    Theorem example (a b : nat) (h : a = b + 1) : a > b.
    ```

    then your final proof snippet should follow this format:

    ```Rocq
    From Stdlib Require Import Lia.

    Theorem example (a b : nat) (h : a = b + 1) : a > b.
    Proof.
      rewrite h.
      lia.
    Qed.
    ```
    The last `Rocq` code block will be parsed as your final proof.
  contribute: [logger, context]
\end{lstlisting}
\caption{\label{code:proof_agent1}PPDL program for MiniF2F benchmark (Part 1: initialization and prompt).}
\end{figure}

\begin{figure}[h]
\begin{lstlisting}[language=pdl,breaklines=true]
- repeat:
    text:
    - text: |
        ╔════════════════════════════════════════════════╗
        ║ [${pdl_particle_id}] ITERATION: ${i}           ║
        ╚════════════════════════════════════════════════╝
      contribute: [logger, stdout, stderr]
    - model: ${ model_params["model_id"] }
      parameters: ${ model_params }
      def: reasoning
      contribute: [context, result, logger]
    - defs:
        response: 
          data: ${ reasoning }
          parser:
            regex:
              (.|\n)*```(coq|Coq|rocq|Rocq|)\n(?P<rocq>(.|\n)*?)```(.|\n)*
            spec: { rocq: string }
      if: ${ response == None }
      then:
        defs:
          response: 
            data: 
              rocq: ${ formal_statement }
    - defs:
        result:
          lang: python
          code: result = checker.check(response["rocq"])
        prompt: 
          match: ${ result["status"] }
          with: 
          - case: Error
            then:
              lastOf: 
              - text: |

                  PROOF ERROR - Attempting to fix ${name}...
                contribute: [logger, stdout]
              - factor: -1
              - text: |
                  Your previous Rocq/Coq proof failed to compile with these verifier
                  errors:

                  ${result["error"]}
                  
                  Please fix your proof.
                contribute: [logger, result]
        tokens: 
          lang: python
          code: |
            result = pdl_usage.completion_tokens + pdl_usage.prompt_tokens
      text: ${ prompt }
  until: ${ (result["status"] == "Complete") or (tokens > agent_params["max_tokens"])} 
  maxIterations: ${ agent_params["max_depth"] }
  index: i
\end{lstlisting}
\caption{\label{code:proof_agent2}PPDL program for MiniF2F benchmark (Part 2: proof/repair loop).}
\end{figure}

\begin{figure}[h]
\begin{lstlisting}[language=pdl,breaklines=true]
- text: |

    Double-checking proof correctness for ${name}...
    Current status: ${ result["status"] }

  contribute: [logger, stdout]
- defs:
    verification: 
      lang: python
      code: result = checker.verify_proof(name, formal_statement, response["rocq"])
- text: |

    ╔════════════════════════════════════════════════╗
    ║ PROOF ${ "COMPLETE" if verification["status"] == "Complete" else "FAILED. " }  ║
    ╚════════════════════════════════════════════════╝
    Status: ${verification["status"]}
    Goal: ${ formal_statement }

    Proof:
    ```rocq
    ${ response["rocq"] }
    ```

  contribute: [logger, stdout]
- defs:
    result:
      data:
        status: ${ verification["status"] }
        proof: ${ response["rocq"] }
        steps: ${ i + 1 }
  lang: python
  code: |
    if (verification["status"] == "Complete"):
      results.append((result, pdl_usage))
      raise Exception("success")
  contribute: [result, logger]

\end{lstlisting}
\caption{\label{code:proof_agent3}PPDL program for MiniF2F benchmark (Part 3: verification and result collection).}
\end{figure}

\section{Case Study: Theorem Proving in Rocq}

We presents additional results for the case study.

\begin{figure}[t]
\centering
\resizebox{0.925\columnwidth}{!}{\definecolor{plotlyblue}{RGB}{31,119,180}
\definecolor{plotlyorange}{RGB}{255,127,14}
\definecolor{plotlygreen}{RGB}{44,160,44}
\definecolor{plotlyred}{RGB}{214,39,40}
\definecolor{plotlypurple}{RGB}{148,103,189}
\definecolor{plotlybrown}{RGB}{140,86,75}
\definecolor{plotlypink}{RGB}{227,119,194}
\definecolor{plotlygray}{RGB}{127,127,127}
\definecolor{plotlyolive}{RGB}{188,189,34}
\definecolor{plotlycyan}{RGB}{23,190,207}
\begin{tikzpicture}
\begin{axis}[
    xlabel={Number of Steps},
    ylabel={Number of Theorems Solved},
    legend pos=south east,
    legend reversed=true,
    legend cell align={left},
    grid=major,
    width=12cm,
    height=8cm,
    line width=1.5pt,
    mark size=2pt
]

\addplot[color=plotlyblue, fill=plotlyblue, fill opacity=0.15, draw=none, forget plot] coordinates {
    (0,0.0)
    (1,43.41421356237309)
    (2,49.80473785412437)
    (3,51.1380711874577)
    (4,52.1380711874577)
    (5,54.580552462257984)
    (6,59.72147133432299)
    (7,60.38813800098966)
    (8,62.16110492451596)
    (9,62.38813800098966)
    (10,62.63299316185545)
    (11,63.36633983786426)
    (12,64.38813800098966)
    (13,65.94392028877594)
    (14,66.96649831220388)
    (15,66.94392028877594)
    (16,67.16110492451597)
    (17,67.16024689946929)
    (18,68.16024689946929)
    (19,68.21895141649746)
    (21,68.36633983786426)
    (23,69.9580026246706)
    (24,70.44948974278317)
    (25,70.38813800098966)
    (27,71.44948974278317)
    (32,72.2007750890142)
    (33,72.53410842234754)
    (35,72.44948974278317)
    (37,72.82777159118262)
    (39,73.82777159118262)
    (44,74.53410842234754)
    (46,75.2659863237109)
    (51,75.2007750890142)
    (53,75.53410842234754)
    (72,75.53410842234754)
    (73,75.53410842234754)
    (74,75.53410842234754)
    (74,69.7992249109858)
    (73,69.7992249109858)
    (72,69.7992249109858)
    (53,69.7992249109858)
    (51,69.46589157765246)
    (46,68.7340136762891)
    (44,68.7992249109858)
    (39,68.83889507548403)
    (37,67.83889507548403)
    (35,67.55051025721683)
    (33,66.7992249109858)
    (32,66.46589157765246)
    (27,66.55051025721683)
    (25,66.278528665677)
    (24,65.55051025721683)
    (23,64.70866404199606)
    (21,64.96699349546908)
    (19,64.4477152501692)
    (18,63.839753100530714)
    (17,62.839753100530714)
    (16,62.172228408817375)
    (15,61.05607971122405)
    (14,60.36683502112944)
    (13,60.05607971122405)
    (12,60.27852866567701)
    (11,59.96699349546907)
    (10,59.36700683814455)
    (9,58.27852866567701)
    (8,57.17222840881737)
    (7,56.27852866567701)
    (6,55.61186199901034)
    (5,52.08611420440869)
    (4,51.19526214587563)
    (3,50.19526214587563)
    (2,48.8619288125423)
    (1,40.58578643762691)
    (0,0.0)
};

\addplot[color=plotlyblue, mark=none, line width=1.5pt] coordinates {
    (0,0.0)
    (1,42.0)
    (2,49.333333333333336)
    (3,50.666666666666664)
    (4,51.666666666666664)
    (5,53.333333333333336)
    (6,57.666666666666664)
    (7,58.333333333333336)
    (8,59.666666666666664)
    (9,60.333333333333336)
    (10,61.0)
    (11,61.666666666666664)
    (12,62.333333333333336)
    (13,63.0)
    (14,63.666666666666664)
    (15,64.0)
    (16,64.66666666666667)
    (17,65.0)
    (18,66.0)
    (19,66.33333333333333)
    (21,66.66666666666667)
    (23,67.33333333333333)
    (24,68.0)
    (25,68.33333333333333)
    (27,69.0)
    (32,69.33333333333333)
    (33,69.66666666666667)
    (35,70.0)
    (37,70.33333333333333)
    (39,71.33333333333333)
    (44,71.66666666666667)
    (46,72.0)
    (51,72.33333333333333)
    (53,72.66666666666667)
    (72,72.66666666666667)
    (73,72.66666666666667)
    (74,72.66666666666667)
};
\addlegendentry{IS@1}

\addplot[color=plotlyorange, fill=plotlyorange, fill opacity=0.15, draw=none, forget plot] coordinates {
    (0,0.0)
    (1,53.41421356237309)
    (2,60.816496580927726)
    (3,64.81649658092772)
    (4,66.1380711874577)
    (5,70.03300650453092)
    (6,69.91388579559131)
    (7,71.721471334323)
    (8,71.60947570824874)
    (9,73.1380711874577)
    (10,74.0)
    (11,75.1380711874577)
    (12,77.63299316185545)
    (13,78.63299316185545)
    (14,79.721471334323)
    (15,80.44948974278317)
    (16,80.36633983786426)
    (17,81.16024689946929)
    (18,82.7578728318319)
    (19,84.36101532453152)
    (20,84.43790283299492)
    (21,84.63316497887055)
    (22,85.55902608401044)
    (26,85.73267967572852)
    (28,85.73267967572852)
    (29,86.52527896759675)
    (30,86.52527896759675)
    (30,78.8080543657366)
    (29,78.8080543657366)
    (28,78.93398699093814)
    (26,78.93398699093814)
    (22,78.44097391598956)
    (21,78.0335016877961)
    (20,76.89543050033842)
    (19,76.30565134213514)
    (18,76.57546050150144)
    (17,76.83975310053071)
    (16,76.96699349546908)
    (15,75.55051025721683)
    (14,75.61186199901034)
    (13,75.36700683814455)
    (12,74.36700683814455)
    (11,74.19526214587565)
    (10,74.0)
    (9,72.19526214587565)
    (8,69.7238576250846)
    (7,67.61186199901034)
    (6,67.41944753774203)
    (5,66.63366016213574)
    (4,65.19526214587565)
    (3,63.183503419072274)
    (2,59.183503419072274)
    (1,50.58578643762691)
    (0,0.0)
};

\addplot[color=plotlyorange, mark=none, line width=1.5pt] coordinates {
    (0,0.0)
    (1,52.0)
    (2,60.0)
    (3,64.0)
    (4,65.66666666666667)
    (5,68.33333333333333)
    (6,68.66666666666667)
    (7,69.66666666666667)
    (8,70.66666666666667)
    (9,72.66666666666667)
    (10,74.0)
    (11,74.66666666666667)
    (12,76.0)
    (13,77.0)
    (14,77.66666666666667)
    (15,78.0)
    (16,78.66666666666667)
    (17,79.0)
    (18,79.66666666666667)
    (19,80.33333333333333)
    (20,80.66666666666667)
    (21,81.33333333333333)
    (22,82.0)
    (26,82.33333333333333)
    (28,82.33333333333333)
    (29,82.66666666666667)
    (30,82.66666666666667)
};
\addlegendentry{IS@5}

\addplot[color=plotlygreen, fill=plotlygreen, fill opacity=0.15, draw=none, forget plot] coordinates {
    (0,0.0)
    (1,61.41421356237309)
    (2,68.41421356237309)
    (3,73.63316497887055)
    (4,74.82842712474618)
    (5,75.41421356237309)
    (6,77.03300650453092)
    (7,77.81649658092772)
    (8,80.38813800098966)
    (9,82.7578728318319)
    (10,82.94392028877594)
    (11,83.94392028877594)
    (12,84.63316497887055)
    (13,86.1919456342634)
    (14,87.10456949966158)
    (15,87.69434865786486)
    (17,88.1919456342634)
    (20,88.1919456342634)
    (21,88.1919456342634)
    (21,80.47472103240325)
    (20,80.47472103240325)
    (17,80.47472103240325)
    (15,79.63898467546848)
    (14,79.56209716700508)
    (13,78.47472103240325)
    (12,78.0335016877961)
    (11,78.05607971122406)
    (10,77.05607971122406)
    (9,76.57546050150144)
    (8,76.278528665677)
    (7,76.18350341907228)
    (6,73.63366016213574)
    (5,72.58578643762691)
    (4,69.17157287525382)
    (3,67.0335016877961)
    (2,65.58578643762691)
    (1,58.58578643762691)
    (0,0.0)
};

\addplot[color=plotlygreen, mark=none, line width=1.5pt] coordinates {
    (0,0.0)
    (1,60.0)
    (2,67.0)
    (3,70.33333333333333)
    (4,72.0)
    (5,74.0)
    (6,75.33333333333333)
    (7,77.0)
    (8,78.33333333333333)
    (9,79.66666666666667)
    (10,80.0)
    (11,81.0)
    (12,81.33333333333333)
    (13,82.33333333333333)
    (14,83.33333333333333)
    (15,83.66666666666667)
    (17,84.33333333333333)
    (20,84.33333333333333)
    (21,84.33333333333333)
};
\addlegendentry{IS@10}

\addplot[color=plotlyred, fill=plotlyred, fill opacity=0.15, draw=none, forget plot] coordinates {
    (0,0.0)
    (1,66.16110492451595)
    (2,73.94392028877594)
    (3,75.41421356237309)
    (4,77.58055246225797)
    (5,80.1380711874577)
    (6,83.58055246225797)
    (7,85.41421356237309)
    (8,86.63299316185545)
    (9,87.58055246225797)
    (10,88.63299316185545)
    (13,88.63299316185545)
    (13,85.36700683814455)
    (10,85.36700683814455)
    (9,85.08611420440869)
    (8,83.36700683814455)
    (7,82.58578643762691)
    (6,81.08611420440869)
    (5,79.19526214587565)
    (4,75.08611420440869)
    (3,72.58578643762691)
    (2,68.05607971122406)
    (1,61.17222840881737)
    (0,0.0)
};

\addplot[color=plotlyred, mark=none, line width=1.5pt] coordinates {
    (0,0.0)
    (1,63.666666666666664)
    (2,71.0)
    (3,74.0)
    (4,76.33333333333333)
    (5,79.66666666666667)
    (6,82.33333333333333)
    (7,84.0)
    (8,85.0)
    (9,86.33333333333333)
    (10,87.0)
    (13,87.0)
};
\addlegendentry{IS@20}

\addplot[color=plotlypurple, fill=plotlypurple, fill opacity=0.15, draw=none, forget plot] coordinates {
    (0,0.0)
    (1,71.41421356237309)
    (2,77.81649658092772)
    (3,82.94392028877594)
    (4,87.02368927062183)
    (5,88.9580026246706)
    (6,89.16110492451597)
    (7,90.02368927062183)
    (8,90.02368927062183)
    (8,85.30964406271151)
    (7,85.30964406271151)
    (6,84.17222840881738)
    (5,83.70866404199606)
    (4,82.30964406271151)
    (3,77.05607971122406)
    (2,76.18350341907228)
    (1,68.58578643762691)
    (0,0.0)
};

\addplot[color=plotlypurple, mark=none, line width=1.5pt] coordinates {
    (0,0.0)
    (1,70.0)
    (2,77.0)
    (3,80.0)
    (4,84.66666666666667)
    (5,86.33333333333333)
    (6,86.66666666666667)
    (7,87.66666666666667)
    (8,87.66666666666667)
};
\addlegendentry{IS@40}

\end{axis}
\end{tikzpicture} }
\caption{Number of MiniF2F-Rocq problems solved as a function of proof/repair steps. Each curve shows the mean over three runs, with shaded regions indicating one standard deviation.}
\label{fig:res_ntp_is_steps}
\end{figure}

\Cref{fig:res_ntp_is_steps} shows the number of problems solved as a function of proof/repair steps for IS depending on the number of particles. It follows the same trends as SMC~(see \Cref{fig:res_ntp_smc_steps}).

\begin{figure}[t]
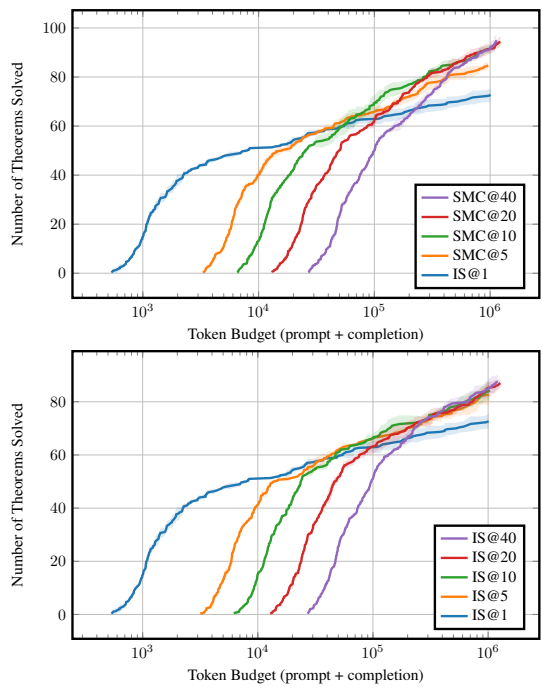

\centering
\resizebox{0.925\columnwidth}{!}{\definecolor{plotlyblue}{RGB}{31,119,180}
\definecolor{plotlyorange}{RGB}{255,127,14}
\definecolor{plotlygreen}{RGB}{44,160,44}
\definecolor{plotlyred}{RGB}{214,39,40}
\definecolor{plotlypurple}{RGB}{148,103,189}
\definecolor{plotlybrown}{RGB}{140,86,75}
\definecolor{plotlypink}{RGB}{227,119,194}
\definecolor{plotlygray}{RGB}{127,127,127}
\definecolor{plotlyolive}{RGB}{188,189,34}
\definecolor{plotlycyan}{RGB}{23,190,207}
 }
\caption{Number of MiniF2F-Rocq problems solved as a function of the number of tokens.}
\label{fig:res_ntp_token}
\end{figure}

\Cref{fig:res_ntp_token} shows the number of MiniF2F-Rocq problems solved as a function of the number of tokens.
IS and SMC consume $k$ times more tokens to prove the first theorem because they run $k$ particles in parallel, but catch up with the IS@1 strategy which can only explore one trajectory.
So the more we increase the token budget, the more the probabilistic approaches are beneficial.

\end{document}